\documentclass{article} 
\usepackage{iclr2027_conference,times}

\usepackage{amsmath,amsfonts,bm}

\def\eqref#1{equation~\ref{#1}}

\def\1{\bm{1}}

\DeclareMathAlphabet{\mathsfit}{\encodingdefault}{\sfdefault}{m}{sl}
\SetMathAlphabet{\mathsfit}{bold}{\encodingdefault}{\sfdefault}{bx}{n}

\usepackage{hyperref}
\usepackage{url}
\usepackage{graphicx}
\usepackage{caption}
\usepackage{booktabs}
\usepackage{float,xcolor}
\usepackage{wrapfig}

\hypersetup{pdftitle={Representations Have Trajectories: Characterizing Temporal Representation Dynamics in Neural Networks}, pdfauthor={Shivaprasad B. Gowda, Sparsh Mittal}}

\title{Representations have Trajectories: Characterizing Temporal Representation Dynamics in Neural Networks}

\author{Shivaprasad B. Gowda \\
Indian Institute of Information Technology, Nagpur \\
\texttt{bt24csa005@iiitn.ac.in}
\And
Sparsh Mittal \\
Indian Institute of Technology, Roorkee \\
\texttt{sparsh.mittal@mfs.iitr.ac.in}
}

\iclrfinalcopy
\begin{document}
\lhead{Preprint}

\maketitle

\begin{abstract}
Recent work has shown that deep neural networks exhibit an implicit low-rank simplicity bias. In this paper, we investigate the questions that naturally arise from this observation. What are the temporal consequences of low-rank simplicity bias during optimization? How does representation geometry evolve throughout optimization rather than only at convergence? Do different neural architectures exhibit distinct geometric optimization behaviors? To answer these questions, we conduct a large-scale empirical study tracking representational geometry throughout training. We find that neural networks follow three recurring geometric trajectories: persistent high-rank, collapse-and-recovery, and persistent low-rank. We further identify Destructive Representation Collapse (DRC) as a pathological regime in which representations collapse to rank-one and fail to recover, leading to a sustained loss of representational capacity. While collapse-and-recovery is consistently associated with successful optimization in controlled settings, experiments on real-world models show that the relationship between geometric trajectories and optimization outcomes is configuration dependent. Overall, our findings suggest that representation geometry evolves dynamically during training and provide a new perspective on how deep neural networks learn.
\end{abstract}

\section{Introduction}

Deep neural networks (DNNs) learn task-relevant representations through nonlinear transformations that evolve throughout optimization \citep{representationlearning2013,saxe2013exact,pmlr-v130-baratin21a}. This evolution, termed \emph{representation dynamics}, offers insight into how networks learn beyond their final representations.

Prior work shows that DNNs exhibit an implicit low-rank bias \citep{huh2023the}, and that under suitable conditions, finite-depth networks can recover the target function's rank \citep{jacot2022implicit}. However, existing studies largely characterize representations at convergence, leaving their evolution during optimization less understood. We therefore ask: \emph{How does representation geometry evolve during optimization, do architectures exhibit distinct trajectories, and are these trajectories associated with optimization success or failure?}

\textbf{Contributions.}
We investigate these questions and make the following observations. In controlled settings, runs can be partitioned into recurring geometric trajectory archetypes. These trajectories include persistent high-rank behavior, transient representation collapse followed by recovery, and persistent low-rank behavior. The prevalence of these trajectories varies considerably across different architectures, optimization settings, and activation functions.
To characterize the pathological extreme of persistent low-rank behavior, we introduce Destructive Representation Collapse (DRC): a persistent rank-one state that, in our experiments, is already present at the first logged epoch and is not escaped during training.
Our main contributions are summarized as follows.

1. A large-scale empirical study of temporal representation geometry during optimization.

2. Identification and characterization of three recurring rank trajectories.

3. Introduction of DRC, with evidence that it is present early, grows with depth, and is rarer with residual connections.

4. Statistical evidence that trajectory types are associated with optimization outcomes at convergence.

5. Evidence that trajectory category at early training milestones forecasts eventual optimization success, replicated across synthetic MLPs.
   
6. CIFAR-10 experiments across MLP, CNN, ResNet, and ViT architectures, replicating the synthetic taxonomy across four architecture families.

\section{Related Works}
\label{sec: related works}
\textbf{Implicit Low-Rank Bias:}
\citet{gunasekar2017implicit} showed that gradient descent in matrix factorization implicitly favors low nuclear-norm solutions, while \citet{arora2019deepmatrixfactorization} showed that depth strengthens this low-rank bias. \citet{huh2023the} found that DNNs favor lower-rank representations and can exploit this bias without increasing capacity. \citet{razin2020implicit} showed that implicit regularization in deep matrix factorization cannot be reduced to simple norm minimization, and \citet{jacot2022implicit} showed that finite-depth networks can recover the target function's bottleneck rank. These works focus primarily on low-rank solutions or the final rank rather than its evolution during training.

\textbf{Representation Geometry:}
\citet{Papyan_2020} showed that deep networks develop structured representations through Neural Collapse, while \citet{pmlr-v130-baratin21a} linked implicit regularization to dynamic alignment with task-relevant directions. \citet{k2024neural} showed that regularized low-rank solutions can achieve lower loss than DNC, while \citet{garrod2026persistence} found that DNC persists under cross-entropy as a stable critical point with a larger optimization basin. \citet{saxe2013exact} showed that deep linear networks follow nonlinear learning trajectories with long plateaus and rapid convergence. Unlike \cite{dohare2024loss}, which studies continual learning degradation, we examine temporal representation dynamics and characterize DRC, a persistent rank-one state associated with optimization failure.

Existing work links DNN implicit bias to low-rank representations, neural collapse, and structured feature geometry, but largely studies final representations or optimization behavior rather than temporal evolution of representation. We address this gap through a layer-wise study of $4{,}084$ controlled synthetic runs, identifying trajectory archetypes missed by endpoint analyses. Representation collapse has also been studied in self-supervised learning \citep{jing2021understanding}, GNN over-smoothing \citep{li2018deeper, oono2019graph}, and Transformer attention collapse \citep{dong2021attention}. Unlike these settings, we study DRC as a persistent rank bottleneck during supervised training associated with optimization failure.

Rank contraction in supervised networks has likewise been linked to depth: \citet{daneshmand2020batch} found stable-rank collapse at initialization in unnormalized networks, prevented by batch normalization, while \citet{feng2022rank} observed decreasing feature and Jacobian rank across layers. Signal-propagation analyses \citep{poole2016exponential, schoenholz2016deep, pennington2017resurrecting} describe ordered and chaotic regimes through the evolution of input correlations; our DRC regime is closely related to the ordered phase, appearing from the first logged epoch and becoming more prevalent with depth. Complementarily, we track full training trajectories, distinguishing runs that escape low-rank states from those that exhibit persistent collapse, and relate these trajectories to optimization outcomes across architectures and optimizers. Although effective rank has been used to assess representation quality and downstream performance \citep{garrido2023rankme}, we use it as a training-time criterion for identifying collapse.

\section{Preliminaries and Experimental Design}
\label{preliminaries_and_experimental_design}
\subsection{Neural Networks and Over-parameterization}
Consider an $L$-layer neural network ($l\in [1,L]$) with input $\mathbf{x} \in \mathbf{R}^{d_0}$. The hidden representations are recursively defined 
\(\mathbf{h}^{(0)}=\mathbf{x},\quad
\mathbf{h}^{(l)}=\sigma\!\left(\mathbf{W}^{(l)}\mathbf{h}^{(l-1)}\right)\) \\
The network output is given by, 
$\mathbf{y}=\mathbf{W}^{(L)}\mathbf{h}^{(L-1)}$,
where $\mathbf{W}^{(l)}$ denotes the weight matrix of the
$l$-th layer and $\sigma(\cdot)$ is a nonlinear activation
function (e.g., ReLU). Setting $\sigma(\cdot)$ to the identity
map yields a linear network. Throughout this work, we consider overparameterized networks in which the number of trainable parameters exceeds the number of training samples. $(N_{samples}=32 \ll N_{parameters})$ with $d_{in}$ and $d_{out}$ = 64, base hidden dimension=256.

\subsection{Teacher-Student Framework}
\label{subsec:teacher_student}
As for synthetic benchmark, we consider a teacher-student regression setting \citep{saxe2013exact} in which the student network learns a target function generated by a teacher matrix $W_{teacher} \in \mathbf{R}^{d_{out} \times d_{in}}$ with controllable rank $r$, where $r \in \{1,\ldots,\min(d_{\mathrm{out}}, d_{\mathrm{in}})\}$. The teacher matrix is constructed as $W_{teacher} = U\Sigma V^{\top}$, where $U \in \mathbf{R}^{d_{out} \times r}$ and $V \in \mathbf{R}^{d_{in} \times r}$ are orthonormal matrices sampled via QR decomposition, $r$ denotes the target rank, and $\Sigma \in \mathbf{R}^{r \times r}$ contains diagonal singular values. Training inputs are sampled independently from standard Gaussian distribution, $\mathbf{x} \sim \mathcal{N}(\mathbf{0}, \mathbf{I})$, and the corresponding targets are generated as $\mathbf{y} = W_{teacher}\mathbf{x} + \epsilon$, where $\epsilon \sim \mathcal{N}(0, \tau^{2}I)$ with $\tau = 0.0$. We employ controlled experiments to isolate the effects of dataset types, task ranks, architecture, initialization, data augmentation, and optimization. Controlled, isolated experiments help us study the representation geometry clearly, without noise artifacts.

\subsection{Representation Geometry}
\label{subsec: representation_geo_definition}
\textbf{Stable Rank.} \citep{rudelson2007sampling} measures how dominant the single strongest direction is relative to the total spectral energy. It is the ratio of the squared Frobenius norm to the squared spectral norm of the activation matrix: \(\mathrm{srank}(\mathbf{H}^{(\ell)})=
\frac{\|\mathbf{H}^{(\ell)}\|_F^2}{\|\mathbf{H}^{(\ell)}\|_2^2}\)
\\
\textbf{Effective Rank.} \citep{effectiverank} measures how many directions carry meaningful mass across the entire spectrum via entropy, capturing representational diversity that becomes hard for stable rank to capture once the tail is small relative to $\sigma_1$. Computed as the exponential of the Shannon entropy of the normalized singular values of the activation matrix:
\(\mathrm{erank}(\mathbf{H}^{(\ell)})=
\exp\!\left(-\sum_i p_i\ln p_i\right)\), where
\(p_i=\frac{\sigma_i}{\sum_j\sigma_j}\). Layer activations \(H^{(\ell)}\) are computed on the $\text{train\_eval}$ loader, an unaugmented, deterministic evaluation subset of the training split; preprocessing follows the released code
\\
We report both the stable and effective ranks because they are sensitive to structurally distinct features of the singular value spectrum. Stable rank is anchored to spectral norm, so it is particularly sensitive to the dominance of a single leading direction, but is comparatively insensitive to how mass is distributed across the remaining spectrum once that mass is small relative to $\sigma_1$. Effective rank is sensitive to how many directions carry non-trivial mass in aggregate, but is less sensitive to extreme dominance by a single direction as long as some distributed tail persists.

\subsection{Geometric Trajectory Analysis}
Stable and Effective ranks are computed across all layers of a network and at specific epochs, such as $\{1, 10, 50, 100, 200,\ldots, T\}$. Ranks are tracked at two different critical locations of the network. (1) \textbf{Min-layer: } The minimum rank observed across all layers at a given epoch. This represents the primary dimensional bottleneck, since it is the narrowest path of the representational pipeline. It is our primary metric for synthetic runs. (2) \textbf{Penultimate-layer: } The rank observed at the second-to-last hidden layer. It is reported to be compatible with prior endpoint-based studies of representational collapse \citep{huh2023the}.

\textbf{Dataset: }As for \textbf{Data Generation}, our input  $(\mathbf{X})$ is a high-dimensional white noise: $\mathbf{X} \sim \mathcal{N}(0,I) \in \mathbf{R}^{\mathrm{N}\times64}$ (one sample per row) and the target is $\mathbf{Y}=\mathbf{X}W^T_{\text{teacher}}$ where $\mathbf{X} \in \mathbf{R}^{\mathrm{N}\times64}$.

Our \textbf{experimental setup in sweeps} is as follows. The input/output dimension $(d_{in},d_{out})$ is constant at $64$. The Task Rank $(r)$ is swept across $\{1,4,16,24,32,64\}$. The training sample size $(N_{train})$ is held constant at $32$ to keep the model in an over-parameterized regime. 
Our \textbf{data splits and sizes} are as follows. The train split is $32$ samples. The validation split is $max(64, \mathrm{N}_{train})$. The test split is $max(256, \mathrm{N}_{train})$. We use a seed value of $42$ for regression and cross-validation analysis.
Information regarding \textit{Architectures}, \textit{Optimizers}, \textit{Initialization schemes}, \textit{Activation functions}, \textit{Residual Connections} used across sweeps can be found in the Appendix Sec.~\ref{sec:architectural specifications}.

\section{Discovery of Geometric Trajectory Archetypes}
\begin{figure}[htbp]
    \centering
    \includegraphics[width=\textwidth]{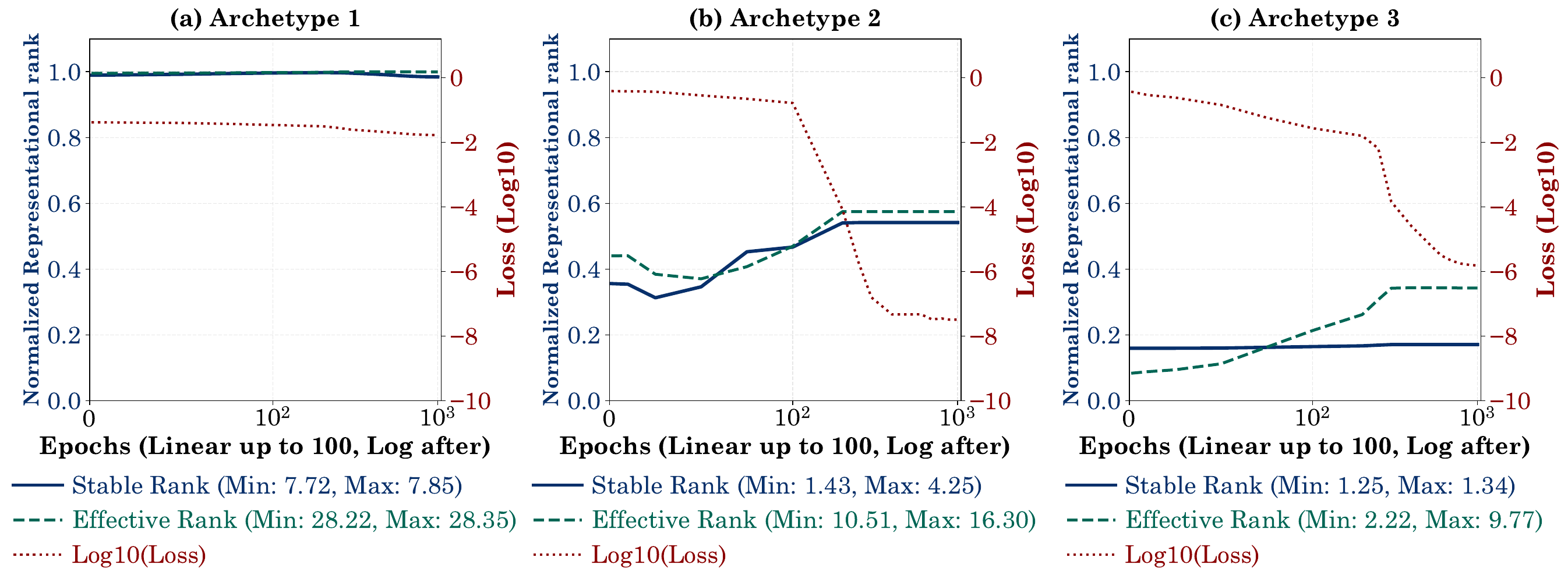}
    \caption{\textbf{Archetypal representation geometry trajectories:} (a) persistent high rank, (b) collapse-and-recovery, and (c) persistent low rank. The x-axis uses a symlog scale to resolve early dynamics.
}
    \label{fig:archetypes_intro}
\end{figure}

Tracking representation geometry throughout optimization shows that runs can be partitioned into recurring geometric trajectory archetypes according to the operational criteria in Table~\ref{tab:archetypes}. These trajectories differ in how the representation geometry evolves over the course of optimization. Figure~\ref{fig:archetypes_intro} visualizes examples.

\noindent\textbf{Archetype 1 (A1): Persistent High Rank.} Throughout optimization, both stable and effective rank remain consistently high with minor fluctuations, indicating that representations preserve their high-dimensional geometry throughout training.

\noindent\textbf{Archetype 2 (A2): Collapse and Recovery.} A2 runs are those in which the stable rank reaches or drops below the collapse threshold of $1.5$ at some logged epoch and later rises above it by more than $0.1$. In the example shown in Figure~\ref{fig:archetypes_intro}(b), this recovery is accompanied by a clear drop in loss. A2 behavior occurs under both default and low-rank initialization (Table ~\ref{tab:initialization_regimes}).

\noindent\textbf{Archetype 3 (A3): Persistent Low Rank.} Representations enter a low-rank regime early in optimization and remain there for the rest of training. Unlike Archetype 2, little or no recovery in representational rank is observed afterward.

\noindent\textbf{DRC: Destructive Representation Collapse.} Within the persistent low-rank regime, we identify an extreme case in which representations are effectively one-dimensional and remain so. Because stable rank alone cannot distinguish moderately low-rank representations from complete collapse, DRC is defined jointly using stable and effective rank (Eq.~\ref{eq:drc_definition}). In our experiments, most DRC runs are already at SR $\le 1.05$ at the first logged epoch (Section~\ref{sec:drc_indepth}), so DRC is better understood as a failure to escape an early rank-one state than as a collapse that develops later in training. 
\begin{table}[ht]
\centering
\caption{\textbf{Rank dynamics archetypes}. Here, $s_t$ denotes stable rank at epoch $t$, $s_{\min}=$ min over logged epochs {$t \in {1,10,50,100,\dots, T}$} of $s_t$. The same notation holds for effective rank $e_t$ as well.}
\label{tab:archetypes}
\small
\resizebox{\columnwidth}{!}{
\begin{tabular}{cl}
\hline
\textbf{Archetype} & \textbf{Criterion} \\
\hline

\textbf{A1} (Persistent High-Rank) &
$\text{SR}_t > 1.5$  at every logged epoch t \\\hline

\textbf{A2} (Transient Collapse and Recovery) &
$(\text{SR}_{\min} \le 1.5) \;\land\; (\text{SR}_{\mathrm{final}} > 1.5) \;\land\; (\text{SR}_{\mathrm{final}} - \text{SR}_{\min} > 0.1)$ \\

\hline

\textbf{A3} (Persistent Low-Rank) &
$(\text{SR}_{\min} \le 1.5) \;\land\; \left[ \text{SR}_{\mathrm{final}} \le 1.5 \;\lor\; (\text{SR}_{\mathrm{final}} - \text{SR}_{\min} \le 0.1) \right] \;\land\;$ \\
&
$(\text{ER}_{\mathrm{final}} > 1.5)$ \\

\hline
\textbf{DRC} (Destructive Representation Collapse) &
$(\text{SR}_{\min} \le 1.05) \;\land\; (\text{SR}_{\mathrm{final}} \le 1.05) \;\land\;$ $(\text{ER}_{\mathrm{final}} \le 1.5)$ \\

\hline
\end{tabular}
}
\end{table}

\noindent Together, these archetypes provide a temporal view of representation dynamics during optimization. Across the $4{,}084$ runs, A1 was the most common trajectory ($1{,}746$, $42.75\%$), followed by DRC ($1{,}106$, $27.08\%$), A3 ($1{,}045$, $25.59\%$), and A2 ($187$, $4.58\%$). A2 had the lowest mean and median losses ($0.0580$ and $0.00001$), while DRC had a median loss of $0.3676$ and a mean loss of $1.05\times10^{15}$ due to divergence. The corresponding mean stable ranks were $7.8135$, $2.7860$, $1.1748$, and $1.00$ for A1--A3 and DRC, respectively. This taxonomy 
captures differences not only in how representation geometry evolves, but also in the optimization outcomes associated with these trajectories.

The observed trajectory patterns are also stable across different initialization seeds. We repeated the full $2,268$-configuration factorial grid using two additional seeds $(43$ and $44)$, resulting in $6,804$ fully trained runs across all three seeds. The DRC fraction remained stable across seeds $(41.80\%, 41.23\%,$ and $42.46\%)$. It differs from the $27.08\%$ reported above, which is over the full 4,084-run pool that also contains the residual sweeps. The association between recovery trajectories and optimization success was also consistent across all three seeds, indicating that the reported geometric dynamics are not specific to the original initialization. (See Table~\ref{tab:appendix_multiseed} for complete details).




\section{Statistical Significance of Geometric Trajectories}

We next examine whether the discovered trajectory archetypes are associated with optimization outcomes. Across all convergence thresholds, Collapse-and-Recovery (A2) trajectories consistently show the highest success rates, whereas DRC exhibits the lowest.
At the strict $10^{-3}$ threshold, optimization success reaches $82.89\%$ for A2 but only $0.90\%$ for DRC; A1 and A3 lie between these extremes at $35.11\%$ and $40.96\%$, respectively. 

 At this threshold, recovery trajectories are $3.07\times$ more likely to optimize successfully than non-recovery trajectories. We tested this association using Fisher's exact test \citep{fisher1922mathematical} and examined whether it persists after controlling for architectural and optimization factors. Archetypes are defined from the trajectory, including its final epoch, so these are associations rather than prospective effects (Table~\ref{tab:contingency_thresholds}).

\begin{table*}[ht]
\centering
\caption{Association between the recovery trajectory (A2) and optimization success at three loss thresholds ($M_\mathrm{models}=4{,}084$, $\text{seed } 42$). Contingency counts, RR (Relative Risk) and the unadjusted OR (Odd's Ratio) compare A2 with all other runs (A1+A3+DRC); the unadjusted OR uses the Haldane–Anscombe correction ((a+0.5)(d+0.5))/((b+0.5)(c+0.5)) with Fisher's exact p. The adjusted OR is the A2 coefficient of the multivariate logistic regression (adjusting for activation, architecture, depth, task rank and optimizer; not initialization) and therefore compares A2 with A1 only. a: A2+success; b: A2+failure; c: non-A2+success; d: non-A2+failure.}

\label{tab:contingency_thresholds}

\setlength{\tabcolsep}{4pt}
\renewcommand{\arraystretch}{0.95}

\resizebox{0.97\textwidth}{!}{
\begin{tabular}{l c c c}
\toprule
\textbf{Metric} &
\textbf{$\text{Loss} < 0.05$} &
\textbf{$\text{Loss} < 10^{-2}$} &
\textbf{$\text{Loss} < 10^{-3}$} \\
\midrule

\textbf{Contingency $(a,b,c,d)$}
& $(164, 23, 1517, 2380)$
& $(156, 31, 1132, 2765)$
& $(155, 32, 1051, 2846)$ \\

\textbf{Success Rate (A2 vs. No A2)}
& $87.70\%$ vs. $38.93\%$
& $83.42\%$ vs. $29.05\%$
& $82.89\%$ vs. $26.97\%$ \\

\textbf{RR [95\% CI]}
& $2.25$ [$2.11, 2.41$]
& $2.87$ [$2.65, 3.11$]
& $3.07$ [$2.83, 3.34$] \\

\textbf{Unadjusted OR [95\% CI]}
& $10.98$ [$7.09, 17.00$]
& $12.13$ [$8.22, 17.90$]
& $12.95$ [$8.82, 19.03$] \\

\textbf{Fisher $p$-value}
& $1.07 \times 10^{-41}$
& $4.74 \times 10^{-51}$
& $2.67 \times 10^{-54}$ \\

\textbf{Adjusted OR [95\% CI]}
& \textbf{6.77 [$3.92, 11.70$]} $p < 0.0001$
& \textbf{4.41 [$2.40, 8.08$]} $p < 0.0001$
& \textbf{4.59 [$2.39, 8.81$]} $p < 0.0001$ \\

\bottomrule
\end{tabular}
}
\end{table*}

The association is statistically significant (Fisher's exact test,  $p=2.67\times10^{-54}$, unadjusted OR = $12.95$). In a multivariate logistic regression adjusting for architecture, optimizer, activation function, depth and task rank (Table~\ref{tab:factors_multivariate}), A2 runs have $4.59\times$ higher odds of success than A1 runs ($p < 0.0001$), this comparison is against A1, not against the pooled non-recovery group. Trajectory type is therefore associated with success beyond these configuration factors. Because archetypes are defined based on the trajectory, including their final epoch, we do not interpret this as a causal or prospective effect. A separate analysis restricted to information available before convergence (Appendix Sec.~\ref{app:early_trajectory_prediction_of_optimization_outcomes}) supports a genuinely prospective claim, see Sec.~\ref{subsec: multivariate analysis.}. A3 has higher unadjusted success than A1 $(40.96\%$ vs. $35.11\%)$ but lower adjusted odds (OR 0.29).

\section{Factors Influencing Geometric Trajectories}
\label{sec:6 factors influencing}
To understand how architectural and hyperparameter choices influence geometric trajectories, we evaluated $M_\mathrm{models}=4{,}084$ synthetic models across different depths, task ranks, architecture classes, activations and optimizer types.
\subsection{Multivariate analysis.}
\label{subsec: multivariate analysis.}

\begin{wrapfigure}{r}{0.48\textwidth}
    \vspace{-10pt}
    \centering
    \includegraphics[width=\linewidth]{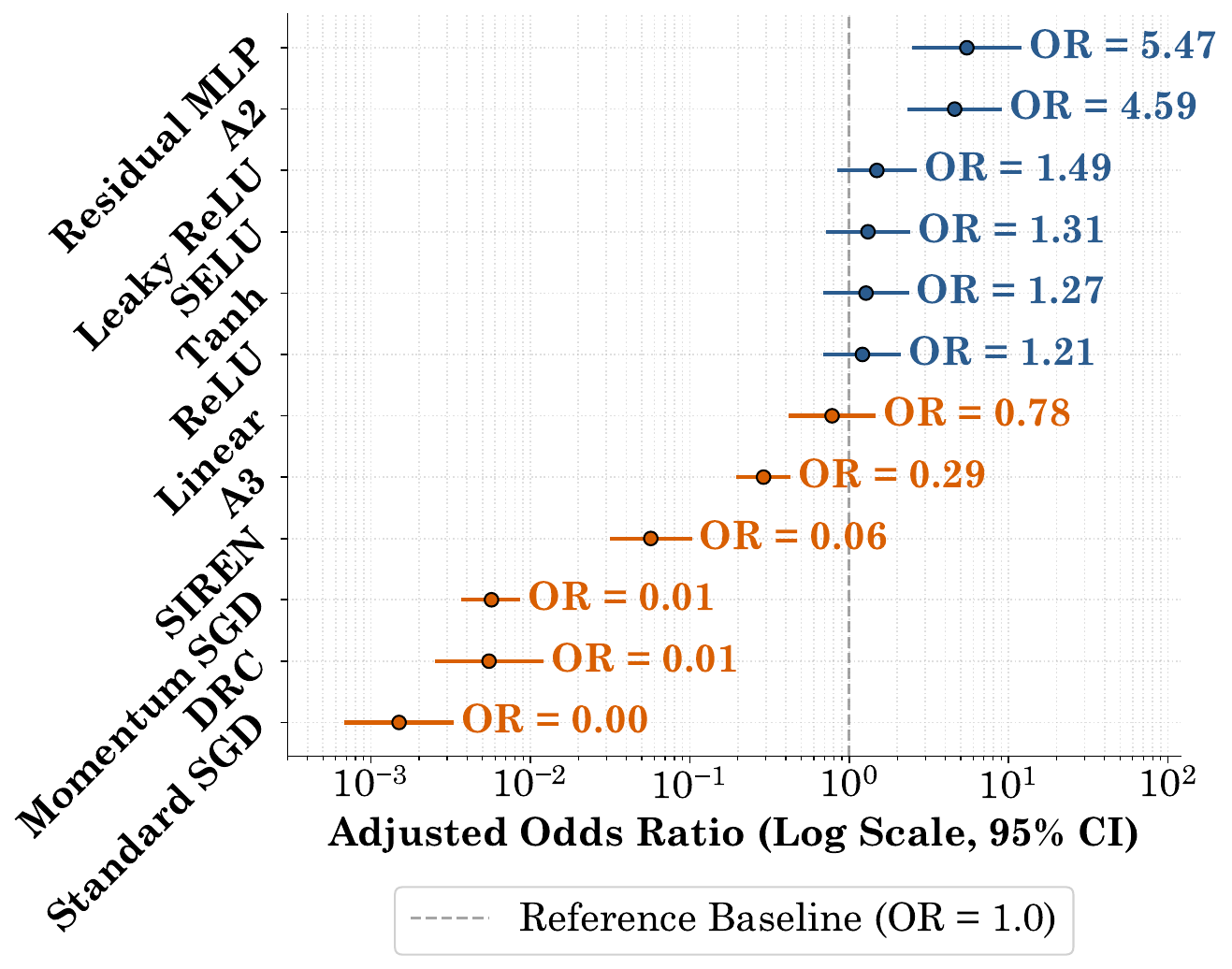}
    \caption{Multivariate Forest Plot: Controlled Predictors of Optimization Success.}
    \label{fig:multivariate_forest}
    \vspace{-10pt}
\end{wrapfigure}

Univariate comparisons can be influenced by confounding factors, as architectural choices (e.g., residual connections or Adam optimization) affect both convergence and representational rank. To isolate the independent effect of each factor on optimization success ($\mathcal{L}_{\text{train}} < 10^{-3}$), we fitted a multivariate logistic regression model including all hyperparameters simultaneously: $\operatorname{logit}(P(Y=1)) = \beta_0 + \sum_{j}\beta_j X_j$. The initialization scheme is not a covariate in this model. From Table~\ref{tab:factors_multivariate} and Fig.~\ref{fig:multivariate_forest}, we note that:

\begin{table}[ht]
\centering
\caption{\textbf{Multivariate logistic regression of factors influencing optimization success ($\mathcal{L}<10^{-3}$, $M_\mathrm{models}=4{,}084$).} Reference baselines: Archetype = A1, Activation = GELU, Architecture = Plain MLP, Optimizer = Adam, Seed = 42, Initialization Scheme is not a covariate. Archetype is defined by the trajectory, so the coefficients of the other factors are conditional on the archetype.}
\label{tab:factors_multivariate}

\fontsize{5.5}{6.5}\selectfont
\setlength{\tabcolsep}{2.5pt}
\renewcommand{\arraystretch}{0.82}

\resizebox{0.88\columnwidth}{!}{
\begin{tabular}{l c c c c}
\toprule
\textbf{Factor / Variable} & \textbf{Coef ($\beta$)} & \textbf{SE} & \textbf{Adjusted OR [95\% CI]} & \textbf{$p$-value} \\
\midrule
Intercept & $+2.0358$ & $0.4662$ & $7.66$ [$3.07, 19.10$] & $<0.0001$ \\
\midrule
Archetype A2 & \textbf{$+1.5229$} & \textbf{$0.3334$} & \textbf{4.59 [$2.39, 8.81$]} & \textbf{$<0.0001$} \\
Archetype A3 & $-1.2377$ & $0.1862$ & 0.29 [$0.20, 0.42$] & $<0.0001$ \\
DRC & \textbf{$-5.2012$} & \textbf{$0.3896$} & \textbf{0.0055 [$0.0026, 0.0118$]} & \textbf{$<10^{-20}$} \\
\midrule
Residual MLP & $+1.6988$ & $0.3917$ & 5.47 [$2.54, 11.78$] & $<0.0001$ \\
Deep Linear MLP & $+0.4347$ & $0.3881$ & 1.54 [$0.72, 3.30$] & $0.2627$ \\
Model Depth ($L$) & $-0.0136$ & $0.0036$ & 0.987 [$0.979, 0.994$] & $0.0002$ \\
Task Rank ($r$) & $-0.0166$ & $0.0036$ & 0.984 [$0.977, 0.990$] & $<0.0001$ \\
\midrule
Momentum SGD & $-5.1735$ & $0.1996$ & 0.0057 [$0.0038, 0.0084$] & $<10^{-20}$ \\
Standard SGD & $-6.5178$ & $0.3954$ & 0.0015 [$0.0007, 0.0032$] & $<10^{-25}$ \\
\midrule
Leaky ReLU & $+0.3987$ & $0.2802$ & 1.49 [$0.86, 2.58$] & $0.1547$ \\
SELU & $+0.2707$ & $0.2982$ & 1.31 [$0.73, 2.35$] & $0.3640$ \\
Tanh & $+0.2423$ & $0.3025$ & 1.27 [$0.70, 2.31$] & $0.4231$ \\
ReLU & $+0.1908$ & $0.2716$ & 1.21 [$0.71, 2.06$] & $0.4824$ \\
Linear & $-0.2478$ & $0.3067$ & 0.78 [$0.43, 1.42$] & $0.4191$ \\
SIREN & $-2.8662$ & $0.2885$ & 0.057 [$0.03, 0.10$] & $<0.0001$ \\
\bottomrule
\end{tabular}
}

{\fontsize{6}{7}\selectfont
\textbf{Coef ($\beta$)} represents the fitted change in log-odds. 
\textbf{SE} measures the coefficient standard error via Hessian inversion. 
\textbf{Adjusted OR} ($e^\beta$) quantifies the factor's independent odds multiplier after controlling for all other covariates. 
\textbf{$p$-value} indicates Wald $z$-test statistical significance.
}

\end{table}

\noindent\textbf{1. Representation Archetypes}: After controlling for architecture, optimizer, activation, depth, and task rank, models following the recovery trajectory (A2) have $4.59\times$ higher odds of successful convergence than the A1 baseline, whereas DRC nearly eliminates the probability of successful optimization (Table~\ref{tab:factors_multivariate}).

\noindent\textbf{2. Architecture Class}: Residual skip connections substantially improve optimization success, increasing the odds of convergence by \textbf{$5.47\times$} relative to Plain MLP baselines. Residual MLPs exhibit a much higher proportion of high-rank trajectories ($\text{A1}=54.50\%$) compared with Plain MLPs ($\text{A1}=32.33\%$) and Deep Linear models ($\text{A1}=21.43\%$), while reducing the occurrence of catastrophic DRC from $43.62\%$ (Plain MLPs) to $8.51\%$ (Residual MLPs).

However, once a model enters a low-rank state ($\text{SR}\le1.5$), recovery remains uncommon across architectures, with similar A2 recovery rates ($5.28\%$ (residual MLPs) vs. $3.80\%$ (plain MLPs)). Instead of eliminating rank decay, residual connections primarily delay its onset to deeper layers, shifting the mean collapse onset at depth $L=64$ from layer $3.81$ in Plain MLPs to layer $16.21$ in Residual MLPs (Fig.~\ref{fig:residual_type_vs_depth_stable_rank}). The residual coefficient remains large after adjusting for archetype (OR = $5.47$), so residual connections likely also help through mechanisms not captured by our rank metrics.

\textbf{3. Optimizer Dynamics}: Within our experimental setting, Adam is associated with substantially higher optimization success than SGD variants. Standard SGD ($\text{Adjusted OR} = 0.0015$, $p < 10^{-25}$) and Momentum SGD ($\text{Adjusted OR} = 0.0057$, $p < 10^{-20}$) reduce optimization success odds by $99.85\%$ and $99.43\%$ relative to Adam.

\textbf{4. Depth ($L$) and Task Rank ($r$)}: Each additional layer of depth reduces success odds by $1.35\%$ ($\text{Adjusted OR} = 0.9865$, $p = 0.0002$).  Each additional unit of teacher rank reduces success odds by about $1.6\%$ ($\text{Adjusted OR} = 0.9836$, $p < 0.0001$)

\textbf{5. Activation Geometry}: Piecewise-linear and smooth monotonic activations (Leaky ReLU, SELU, Tanh, and ReLU) perform similarly to the GELU baseline, with no statistically significant difference in optimization success ($\text{Adjusted OR}=1.21$--$1.49$, $p>0.15$). In contrast, periodic SIREN activations are associated with substantially lower convergence success overall, reducing the odds of successful convergence by $\sim95\%$ after controlling for optimizer effects. The SGD-specific failure of SIREN activations is examined separately in Sec.~\ref{subsec: dead_gradiet_SIREN_effect} (Fig. ~\ref{fig:cifar10_grad_proof}). SIREN runs almost never collapse to rank one, yet rarely succeed, so a high rank is not sufficient for optimization success.

\textbf{Early trajectory forecasting and association with outcomes.} Trajectory category provides a strong early signal of eventual optimization success in synthetic MLPs. 
Using only information available at each milestone, without final-epoch leakage, trajectory categories at $10\%, 25\%$, and $50\%$ of training predict eventual success with odds ratios of $6.33\times$, $10.60\times$, and $12.11\times$, respectively. 
The classification uses only information available up to the respective milestone and does not use final loss or final representations. These effects remain stable across 20 threshold combinations (Appendix Sec.~\ref{app:early_trajectory_prediction_of_optimization_outcomes},~\ref{app:sensitivity_to_trajectory_classification_thresholds}). The $25\%$ milestone therefore provides a particularly clear early indication of eventual optimization success. For completeness, retrospective association analyses using the full trajectory show that adding trajectory archetypes to hyperparameter controls increases AUC from $0.9459$ to $0.9660$ (accuracy $90.77\% \rightarrow 92.58\%$), while final-rank metrics alone achieve an AUC of $0.7743$ ($69.20\%$) (Table~\ref{tab:cv_classifier_performance}).

\section{Destructive Representation Collapse (DRC)}
\label{sec:drc_indepth}

\begin{wrapfigure}{r}{0.48\columnwidth}
    \vspace{-5pt}
    \centering

    \captionsetup{font=footnotesize,skip=2pt}

    \includegraphics[
        width=0.96\linewidth,
        keepaspectratio
    ]{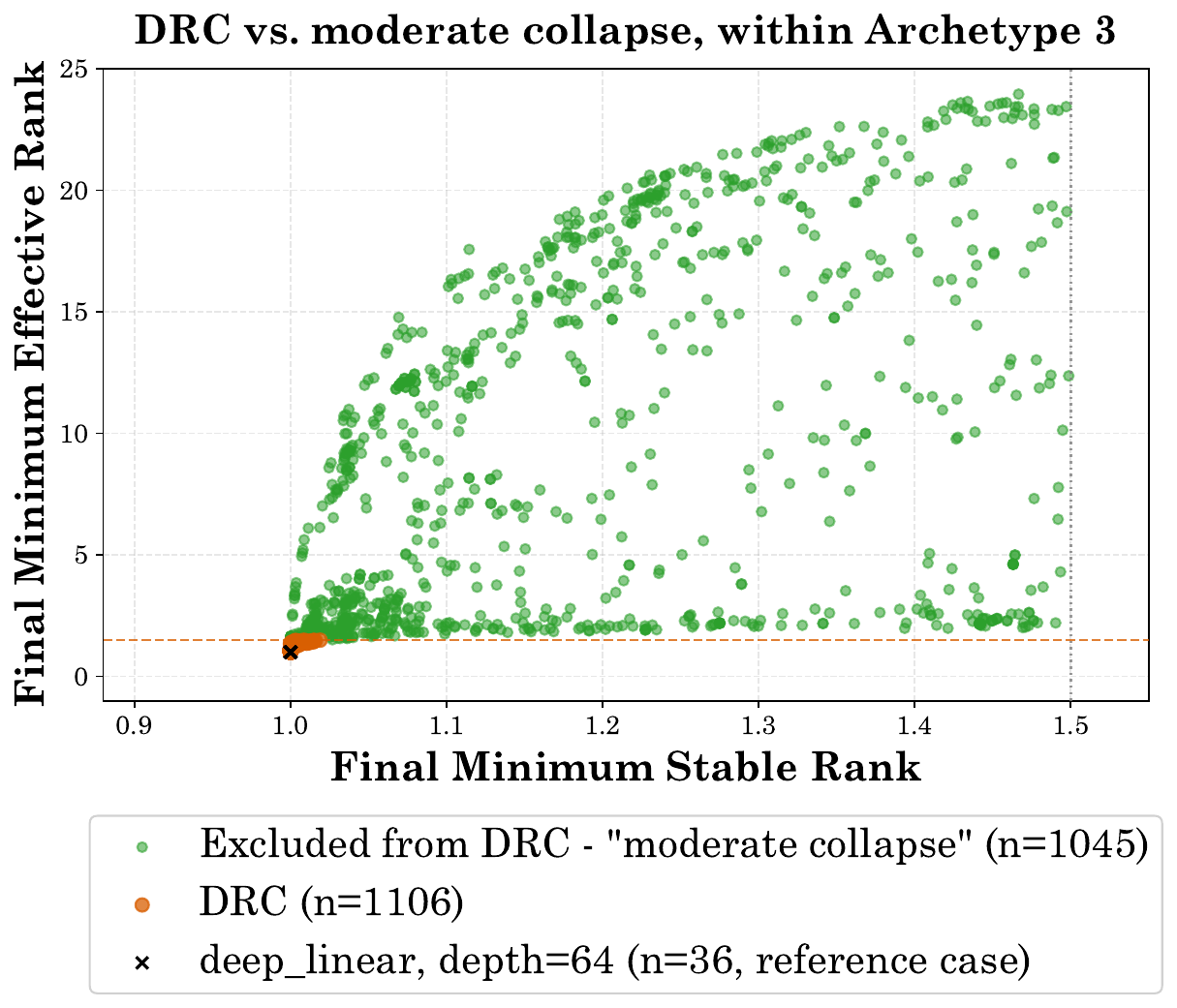}

    \caption{Empirical separation of DRC from moderate low-rank collapse (A3).}
    \label{fig:drc_vs_moderate_collapse}

    \vspace{-8pt}
\end{wrapfigure}
\begin{wrapfigure}{r}{0.48\columnwidth}
    \vspace{-10pt}
    \centering

    \captionsetup{font=footnotesize}

    \includegraphics[
        width=0.96\linewidth,
        keepaspectratio
    ]{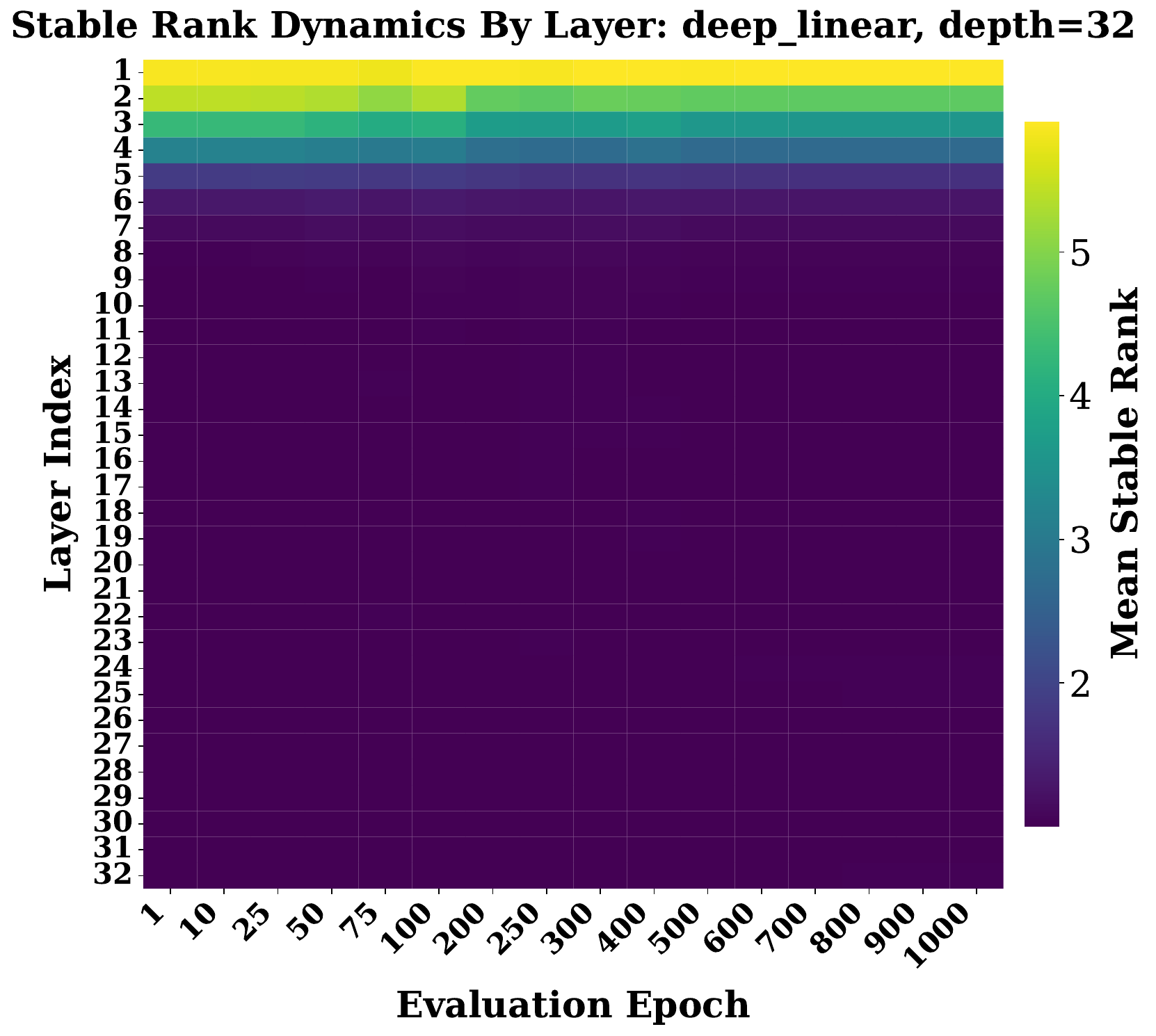}

    \caption{Layer-wise stable rank dynamics: deeper layers undergo DRC without recovery.}
    \label{fig:drc_layerwise}

    \vspace{-10pt}
\end{wrapfigure}

Prior work has shown that deep networks develop low-rank representations and that rank contracts with depth in unnormalized feedforward networks (Sec.~\ref{sec: related works}).
In our synthetic experiments, the most severe form of this contraction, an effectively rank-one representation, becomes more frequent with depth in networks without residual connections or depth-dependent scaling (Fig.~\ref{fig:drc_incidence_rate_depth}), varies with activation function (Fig.~\ref{fig:drc_activation_vulnerability}), and is associated with optimization failure (Figs~\ref{fig:drc_vs_moderate_collapse}-~\ref{fig:drc_layerwise}, Tables~\ref{tab:success_probability}-~\ref{tab:architecture_drc_breakdown}).
We call this state DRC.

Stable rank alone does not identify complete collapse; it is dominated by the leading singular value $\sigma_1$ and is insensitive to the rest of the spectrum. A model can have low stable rank (SR $\approx 1.05–1.19$) while retaining substantial effective rank (ER $\geq 10.4$), as in a Plain CNN run that reaches $90.35\%$ CIFAR-10 test accuracy (Table~\ref{tab:benign_low_rank_milestones}, Fig.~\ref{fig:lowstable_higheffective_goodperformance_Drc}). Hence, we  define DRC using effective rank, together with stable-rank conditions (Eq.~\ref{eq:drc_definition})
A DNN comes under the DRC regime if it satisfies the following criterion: 
\begin{equation}
\colorbox{gray!6}{$
\displaystyle
\bm{
\mathrm{SR}_{\min} \le 1.05
\;\land\;
\mathrm{SR}_{\mathrm{final}} \le 1.05
\;\land\;
\mathrm{ER}_{\mathrm{final}} \le 1.5
}
$}
\label{eq:drc_definition}
\end{equation}

Each component captures a different aspect of collapse. The minimum stable rank indicates whether the representation ever collapses to a nearly one-dimensional state; the final stable rank confirms that it never recovers, and the effective-rank criterion ensures that the remaining singular-value spectrum has also collapsed rather than retaining meaningful high-dimensional structure. Appendix Sec.~\ref{sec: support_analysis_DRC} justifies the selected thresholds to define DRC. Across the synthetic pool, DRC occurs in $1,106$ of $4,084$ runs $(27.08\%)$, but its prevalence varies substantially by architecture: $60.32\%$ of Deep Linear runs, $43.62\%$ of Plain MLP runs, and $8.51\%$ of Residual MLP runs exhibit DRC. Architecture-wise breakdown is reported in Table~\ref{tab:architecture_drc_breakdown}.

Two observations support the association of DRC with optimization failure. 

1. For image classification on CIFAR-10, all $100.0\%$ of models $(41/41)$ that entered the DRC regime collapse to a random guess (Accuracy $\leq 10.15\%$, mean $9.99\%$), with cross-entropy loss freezing at baseline $\approx 2.3026$.

2. In synthetic matrix regression, DRC accounts for $1{,}106$ of $4{,}084$ synthetic runs ($27.08\%$). Crucially, \textbf{$100.0\%$ of all numerical loss explosions} ($152/152$ runs with $\mathcal{L} > 100$) occur exclusively within the DRC regime, with losses expanding up to $10^{15}$. Exploding activations are themselves dominated by a single direction, so these runs are not independent evidence that collapse precedes failure. Among the $954$ non-divergent DRC runs, the vast majority fail to optimize, freezing at high unlearned baseline loss ($\text{median MSE} \approx 0.29$), while only $0.90\%$ of all DRC runs reach \(L<10^{-3}\) (Table~\ref{tab:success_probability})

\textbf{When DRC arises.} In the three-seed grid $(6,804$ runs), all $2,846$ models already have SR $\leq 1.05$ at epoch 1. With full-batch training, epoch 1 is one gradient step from initialization, indicating that DRC is present essentially from the start rather than emerging during training, consistent with signal-propagation failure in deep unnormalized networks (Sec.~\ref{sec: related works}). DRC is less prevalent with default initialization ($29.3\%$) than with the rank-one initialization ($54.4\%$) used in half the grid  (Table~\ref{tab:initialization_regimes}), but remains frequent under default initialization. We therefore interpret DRC as identifying runs trapped in an early rank-one state, not as evidence that training dynamics cause rank collapse.
The probability that a model undergoes DRC scales nearly monotonically with architectural depth $L$. (Fig.~\ref{fig:drc_incidence_rate_depth}). 

Importantly, our results show that low-rank representations should not be viewed as uniformly pathological. Instead, DRC represents an extreme point on the low-rank spectrum that is consistently associated with optimization failure (Fig.~\ref{fig:drc_vs_moderate_collapse}).

\section{CIFAR-10 Experiments}

The preceding sections establish the proposed trajectory taxonomy using controlled matrix regression experiments. To test consistency, we now ask which of these patterns appear in $306$ CIFAR-10 \citep{krizhevsky2009learning} vision models, spanning MLP, CNN, ViTs and residual architectures. 

\subsection{Architecture-Dependent Recovery Trajectories}

Across broader CIFAR-10 architectural sweeps, representational trajectories exhibit the following patterns:

\textbf{Plain MLPs.} As in the controlled experiments, models following the recovery trajectory consistently optimize better than those remaining in the persistent low-rank regime. Recovery is associated with lower training loss ($0.8380$ A2 vs. $2.0072$ A3; $1.1692$ nats reduction) and test loss ($1.5204$ A2 vs. $2.1899$ A3) (Table~\ref{tab:cifar10_cnn_resmlp_mlp}). This is consistent with the synthetic results.

\textbf{Residual MLPs and ResNets.} In the CIFAR-10 sweeps, residual architectures reached low training and test loss and did not enter the DRC regime. The $6$ ResNet runs (ResNet$-18/50 \times 3$ skip modes) retain high effective rank $(11.58–14.45)$ at the end of training. These observations are consistent with skip connections delaying rank decay (Section~\ref{sec:6 factors influencing}).

\textbf{Vision Transformers.} Across 48 CIFAR-10 ViT runs with and without residual connections, no-residual ViTs showed strong depth-wise rank contraction, with 2 of the 8 depth-9 no-residual runs collapsing under the dual-metric DRC criterion to rank one at chance accuracy, whereas zero residual ViTs collapsed (Table~\ref{tab:depth-wise-scaling}). Early Recovery runs showed substantially higher eventual success than Early Stuck runs, with odds ratios of $23.00\times$, $12.78\times$, and $12.78\times$ at $10\%, 25\%,$ and $50\%$ of training, respectively (Table~\ref{tab:early-trajectory-archetype}), indicating the geometric framework extends to Vision Transformers; full details are in Appendix Sec.~\ref{app:extension to Vision Transformers}.

\begin{figure}[htbp]
\centering
\includegraphics[width=0.95\columnwidth,height=0.8\textheight,
keepaspectratio]{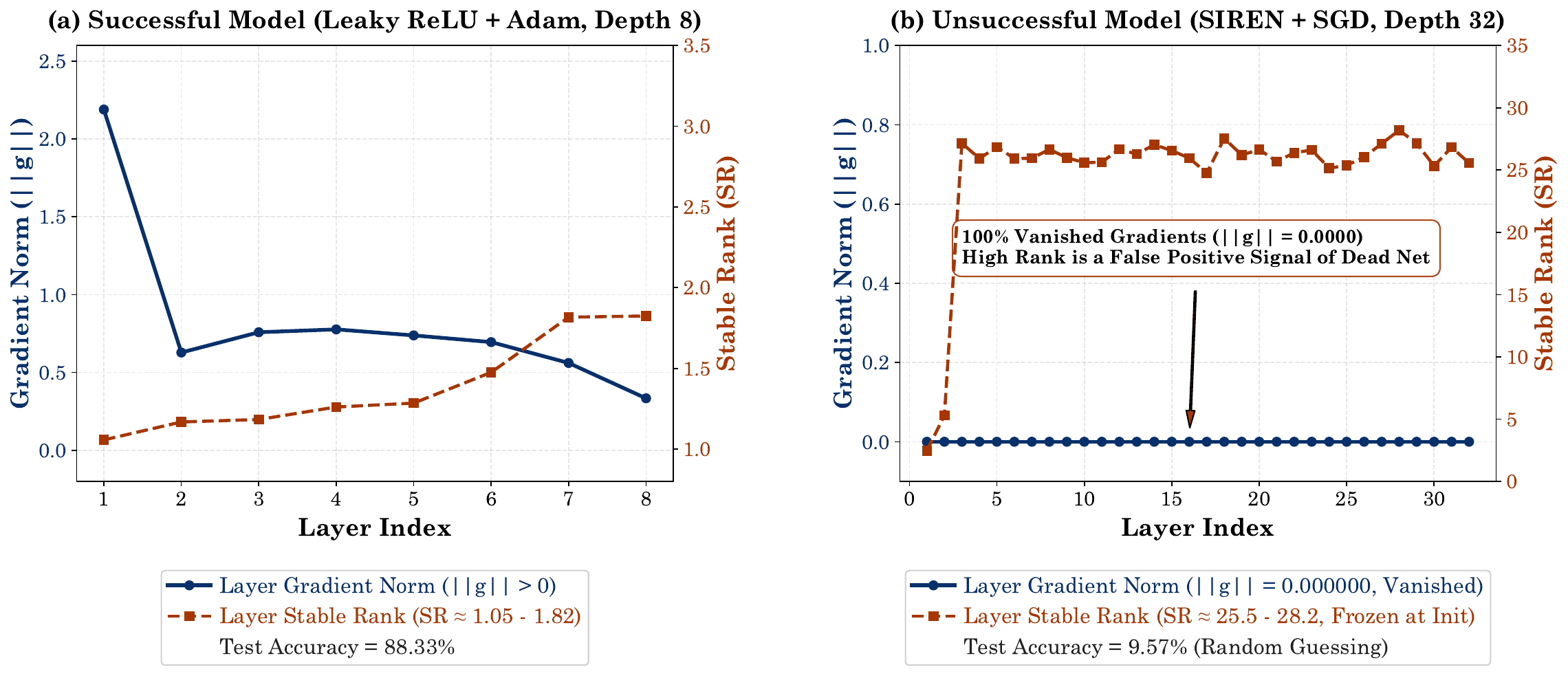}
\captionsetup{font=footnotesize,skip=2pt}
\captionof{figure}{ Active learning in a successful model (A) versus vanished layerwise gradients ($\|g\|=0.0000$) in an un-converged model (B).}
\label{fig:cifar10_grad_proof}
\end{figure}

\subsection{Dead gradient effect in Plain Deep CNNs.}
\label{subsec: dead_gradiet_SIREN_effect}
In plain feedforward CNNs ($M_\mathrm{models} = 192$), a superficial examination suggests an apparent contradiction: models achieving low cross-entropy loss predominantly reside in the low-rank regime ($\text{SR}_{\text{final}} \le 1.5$), whereas high-rank networks exhibit poor loss optimization. Tracking layerwise gradient norms ($\|g_l\|$) resolves this issue:

\noindent\textbf{1. Dead-Gradient High-Rank scenario}:  $9.38\%$ of plain CNN runs ($18 / 192$) remain in Archetype 1. Crucially, $100.0\%$ of these 18 high-rank models failed to optimize, freezing at high test cross-entropy loss ($\text{mean} = 2.6445$, range: $2.3353 - 2.7659$) and high training loss ($\text{mean} = 2.6550$). As illustrated in Figure~\ref{fig:cifar10_grad_proof}(b), these runs consist exclusively of deep ($L=32$) SIREN-activation CNNs under SGD whose layerwise gradient norms vanished to absolute zero ($\|g_l\| = 0.000000, \forall l \in [1, 32]$). In this regime, a high stable rank does not indicate representational capacity in use; it reflects a network that has not been trained.

\noindent\textbf{2. Low-Rank Regularized models}: Excluding unconverged dead gradient runs, other plain CNN models operate in Archetype 2-3 and span a wide range of test accuracy (Figure~\ref{fig:cifar10_grad_proof}(a)).

\noindent \textbf{3. Destructive Collapse (DRC)}: In contrast, $21.35\%$ of plain CNN runs ($41 / 192$) satisfy the DRC criterion, all 41 are at chance accuracy with test cross-entropy near the random guess baseline $(-\ln 0.1 \approx 2.3026,$ mean $2.4445)$.

 These findings align with Section~\ref{sec:drc_indepth}: low-rank representations are not inherently pathological, but the rank-one DRC regime and high-rank runs with vanished gradients are both associated with failure. Rank metrics, therefore, complement rather than replace loss curves and gradient diagnostics. Since DRC is visible at the first logged epoch, monitoring SR and ER could help flag doomed runs early, although we have not tested this as an early-stopping rule.

\section{Conclusion \& Future Work}
\label{sec:discussion_conclusion}

Overall, representation trajectories reveal optimization behavior beyond static endpoints: successful runs tend to avoid persistent Destructive Representation Collapse (DRC), while recovery from low-rank states is associated with convergence. Residual connections are associated with fewer DRC cases and delayed rank decay, and trajectory categories at $25\%$ of training already forecast eventual outcomes, suggesting potential prospective value.

Future work includes formal bounds, comparisons with Hessian, edge-of-stability, and plasticity-loss diagnostics (e.g., dormant neurons), activation-centering and stabilization in ViTs, larger per-architecture CIFAR-10 sweeps, higher-order spectral measures such as Schatten norms and singular-value decay, and evaluating early-milestone trajectory categories as practical early-stopping signals.

\subsubsection*{AI and LLM Disclosure Statement}
  In the interest of transparency, we disclose that large language models (specifically OpenAI ChatGPT, Google Gemini and Anthropic Claude along with Research engines such as Perplexity and Gemini Deep search) were utilized during the preparation of this manuscript under strict authorial oversight for the following specified tasks:
\begin{itemize}
    \item \textbf{Writing Assistance and Editorial Polishing:} LLMs were employed to improve grammatical precision, sentence clarity, academic tone, and typographical flow, as well as to assist with formatting complex \LaTeX{} tables and multi-panel figures.
    \item \textbf{Literature Retrieval and Discovery:} LLMs were queried to assist in identifying and indexing relevant prior literature across dimensional collapse, matrix factorization dynamics, neural network initialization theory, and transformer representation scaling.
    \item \textbf{Coding and Experimental Assistance:} LLMs were used to assist with writing, debugging, and refining experimental code, including data processing, model implementation, training pipelines, and analysis scripts. All generated or suggested code was reviewed, modified, and validated by the authors before use.
\end{itemize}
\textbf{Scope of Human Verification and Exclusions:} Large language models were \emph{not} used to generate experimental data, simulate run logs, formulate mathematical proofs, or synthesize scientific conclusions. All empirical runs across multi-seed grids, CIFAR-10 experiments, Singular Value Decompositions (SVD), effective rank/stable rank evaluations, and statistical hypothesis tests were executed on local computational hardware and computed directly from empirical logs. All cited publications, historical attributions, mathematical definitions, and empirical interpretations were verified and cross-checked manually by the authors. The authors retain full accountability and responsibility for the entire scientific content, claims, and integrity of this manuscript.

\bibliography{iclr2027_conference}
\bibliographystyle{iclr2027_conference}
\clearpage
\renewcommand\thetable{S.\arabic{table}} 
\setcounter{table}{0}  
 
\renewcommand\thefigure{S.\arabic{figure}} 
\setcounter{figure}{0}

\appendix

\section*{Appendix}
\addcontentsline{toc}{section}{Appendix}
\section{Trajectory Archetypes and Optimization Outcomes}
\begin{table}[H]
\centering
\small
\caption{\textbf{Trajectory archetype distribution and optimization success across three random initialization seeds.}
Each seed evaluates the same $N=2{,}268$ configurations, yielding $6{,}804$ fully trained runs in total.
Success is defined as $\mathcal{L}_{\mathrm{train}} < 0.01$.}
\label{tab:appendix_multiseed}
\begin{tabular}{cccccccc}
\toprule
\textbf{Seed} & \textbf{A1 (\%)} & \textbf{A2 (\%)} & \textbf{A3 (\%)} & \textbf{DRC (\%)} & \textbf{A2 Succ. (\%)} & \textbf{A3 Succ. (\%)} & \textbf{DRC Succ. (\%)} \\
\midrule
\textbf{42} & $31.70\%$ & $5.42\%$ & $21.08\%$ & $41.80\%$ & $80.49\%$ & $32.43\%$ & $1.48\%$ \\
\textbf{43} & $33.55\%$ & $4.72\%$ & $20.50\%$ & $41.23\%$ & $88.79\%$ & $30.97\%$ & $1.28\%$ \\
\textbf{44} & $33.99\%$ & $5.34\%$ & $18.21\%$ & $42.46\%$ & $81.82\%$ & $34.87\%$ & $1.56\%$ \\
\midrule
\textbf{Mean $\pm$ SD} & $33.08\pm1.22\%$ & $5.16\pm0.38\%$ & $19.93\pm1.52\%$ & $41.83\pm0.62\%$ & $83.70\pm4.46\%$ & $32.75\pm1.97\%$ & $1.44\pm0.14\%$ \\
\bottomrule
\end{tabular}
\end{table}

\begin{table}[H]
\centering
\caption{\textbf{Summary statistics across trajectory archetypes.}}
\label{tab:trajectory_summary}

\begin{tabular}{lccccc}
\hline
\textbf{Type} &
\textbf{$M_\mathrm{models}$} &
\textbf{\%} &
\textbf{Mean Loss} &
\textbf{Median Loss} &
\textbf{Mean SR} \\
\hline

A1  & 1,746 & 42.75 & 0.7127 & 0.0887   & 7.8135 \\
A2  &   187 &  4.58 & \textbf{0.0580} & \textbf{0.00001} & 2.7860 \\
A3  & 1,045 & 25.59 & 0.1846 & 0.0223   & 1.1748 \\
DRC & 1,106 & 27.08 & $1.05\times10^{15\dagger}$ & 0.3676 & 1.00 \\

\hline
\end{tabular}

\vspace{1mm}

{\scriptsize
$\dagger$ Diverged during optimization with exploded loss values.
}
\end{table}

\begin{table}[H]
\centering
\caption{\textbf{Success probability conditioned on trajectory archetype.}
The probability of successful optimization varies substantially across A1--A3 and DRC.}
\label{tab:success_probability}

\small
\setlength{\tabcolsep}{4pt}
\renewcommand{\arraystretch}{1.15}

\begin{tabular}{lcc}
\toprule
\textbf{} &
\textbf{$<10^{-2}$} &
\textbf{$<10^{-3}$} \\
\midrule
\textbf{A1} &
37.80\% &
35.11\% \\

\textbf{A2} &
\textbf{83.42\%} &
\textbf{82.89\%} \\

\textbf{A3} &
43.73\% &
40.96\% \\

\textbf{DRC} &
1.36\% &
0.90\% \\
\bottomrule
\end{tabular}
\end{table}

\begin{table}[H]
\centering

\caption{\textbf{Architecture class-wise DRC breakdown.} Frozen = DRC runs that did not diverge $(L \leq 100)$, this includes runs on low-rank teachers that reach low absolute loss.}
\label{tab:architecture_drc_breakdown}

\small
\setlength{\tabcolsep}{3pt}
\renewcommand{\arraystretch}{0.95}

\resizebox{\textwidth}{!}{
\begin{tabular}{lccccc}
\toprule
\textbf{Architecture} &
\textbf{Total} &
\textbf{DRC} &
\textbf{DRC Rate} &
\textbf{Diverged} &
\textbf{Frozen} \\
\midrule
Deep Linear & 252 & 152 & 60.32\% & 15 & 137 \\
Plain MLP & 1,788 & 780 & 43.62\% & 39 & 741 \\
Residual MLP & 2,044 & 174 & 8.51\% & 98 & 76 \\
\midrule
\textbf{Total Synthetic Pool} &
\textbf{4,084} &
\textbf{1,106} &
\textbf{27.08\%} &
\textbf{152} &
\textbf{954} \\
\bottomrule
\end{tabular}
}

\end{table}

\begin{table}[H]
\centering
\caption{The 2,268-configuration grid is balanced across initialization schemes (1,134 configurations each per seed). Counts are summed over seeds 42, 43, 44.}
\label{tab:initialization_regimes}
\small
\begin{tabular}{lccccc}
\toprule
\textbf{Init} & \textbf{A1} & \textbf{A2} & \textbf{A3} & \textbf{DRC} & \textbf{Total} \\
\midrule
Default (PyTorch Kaiming-uniform; Table~\ref{tab:initialization})
& 1,649 (48.5\%) & 215 (6.3\%) & 542 (15.9\%) & 996 (29.3\%) & 3,402 \\

Low-rank ($r=1$)
& 602 (17.7\%) & 136 (4.0\%) & 814 (23.9\%) & 1,850 (54.4\%) & 3,402 \\

\textbf{Total}
& \textbf{2,251 (33.1\%)} & \textbf{351 (5.2\%)} & \textbf{1,356 (19.9\%)} & \textbf{2,846 (41.8\%)} & \textbf{6,804} \\
\bottomrule
\end{tabular}
\end{table}

\section{Formal Hypothesis Testing for Task-Rank Invariance}

To evaluate whether the target task rank ($r\in\{1,4,16,24,32,64\}$) influences representation collapse, we performed statistical tests over all $N=2{,}268$ synthetic models.

\begin{table}[H]
\begin{center}
\captionof{table}{Statistical Analysis of the Influence of Task Rank on Representation Metrics. SR and ER denote Stable Rank and Effective Rank, respectively.}

\scriptsize
\setlength{\tabcolsep}{1.5pt}

\begin{tabular}{cccccc}
\toprule
\textbf{Target} &
\textbf{Runs} &
\textbf{Mean SR} &
\textbf{Median SR} &
\textbf{Mean ER} &
\textbf{Median ER} \\
\textbf{Rank} &
\textbf{($N$)} & & & & \\
\midrule

$r=1$  & 378 & $2.7229 \pm 3.997$ & 1.0003 & $8.8891 \pm 11.482$ & 1.1900 \\

$r=4$  & 378 & $2.7630 \pm 4.010$ & 1.0004 & $8.9129 \pm 11.451$ & 1.2003 \\

$r=16$ & 378 & $2.8345 \pm 4.110$ & 1.0003 & $8.9942 \pm 11.449$ & 1.2088 \\

$r=24$ & 378 & $2.7768 \pm 3.960$ & 1.0004 & $9.0070 \pm 11.452$ & 1.2312 \\

$r=32$ & 378 & $2.7448 \pm 3.966$ & 1.0004 & $8.9510 \pm 11.414$ & 1.1969 \\

$r=64$ & 378 & $2.8307 \pm 4.108$ & 1.0008 & $8.9635 \pm 11.419$ & 1.2254 \\

\bottomrule
\end{tabular}
\end{center}
\end{table}

\paragraph{Spearman Rank Correlation.}
The null hypothesis ($H_0$) assumes no monotonic relationship between task rank and minimum representation rank. Results indicate no significant correlation for either stable rank ($\rho=0.0076,\; p=0.7162$) or effective rank ($\rho=0.0040,\; p=0.8499$), failing to reject $H_0$.

\begin{figure}
\centering
\includegraphics[width=0.9\columnwidth,height=0.5\textheight,
keepaspectratio]{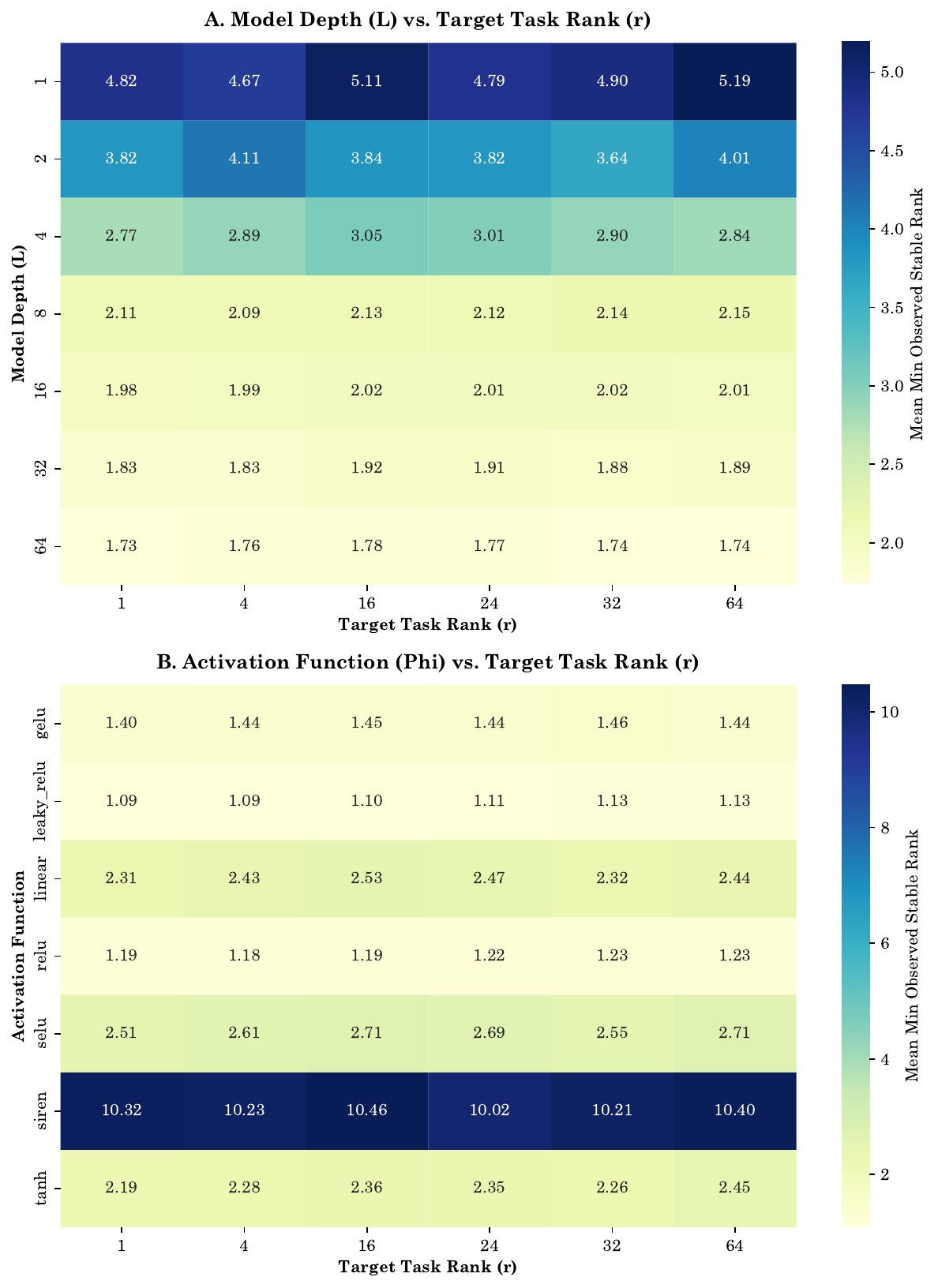}
\captionsetup{font=footnotesize,skip=2pt}
\captionof{figure}{
Mean min-observed representation rank remains uniform across target task ranks $r\in\{1,\ldots,64\}$, confirming that early-layer rank collapse is governed by architectural depth and activation geometry rather than target task rank.
}
\label{fig:task_rank_heatmaps}
\end{figure}

\paragraph{Conclusion.}
These results show no evidence of a monotonic association between target task rank and minimum representation rank. See Fig.~\ref{fig:task_rank_heatmaps}.

\paragraph{}The $1,344$ synthetic runs of Table 16 indicate that representational stability depends strongly on the residual variant. Learnable-gate connections retain the most rank (A1 = $78.21\%$, DRC = $2.14\%$). Unscaled post-activation variants with unbounded activations (ReLU, Leaky ReLU) at high depth$ (L \geq 32)$ are associated with numerical loss divergence rather than slow rank collapse ( Figs.~\ref{fig:residual_type_vs_depth_stable_rank} -~\ref{fig:residual_type_vs_depth_effective_rank}).

\centering
\includegraphics[width=1.2\columnwidth,height=1.5\textheight,
keepaspectratio]{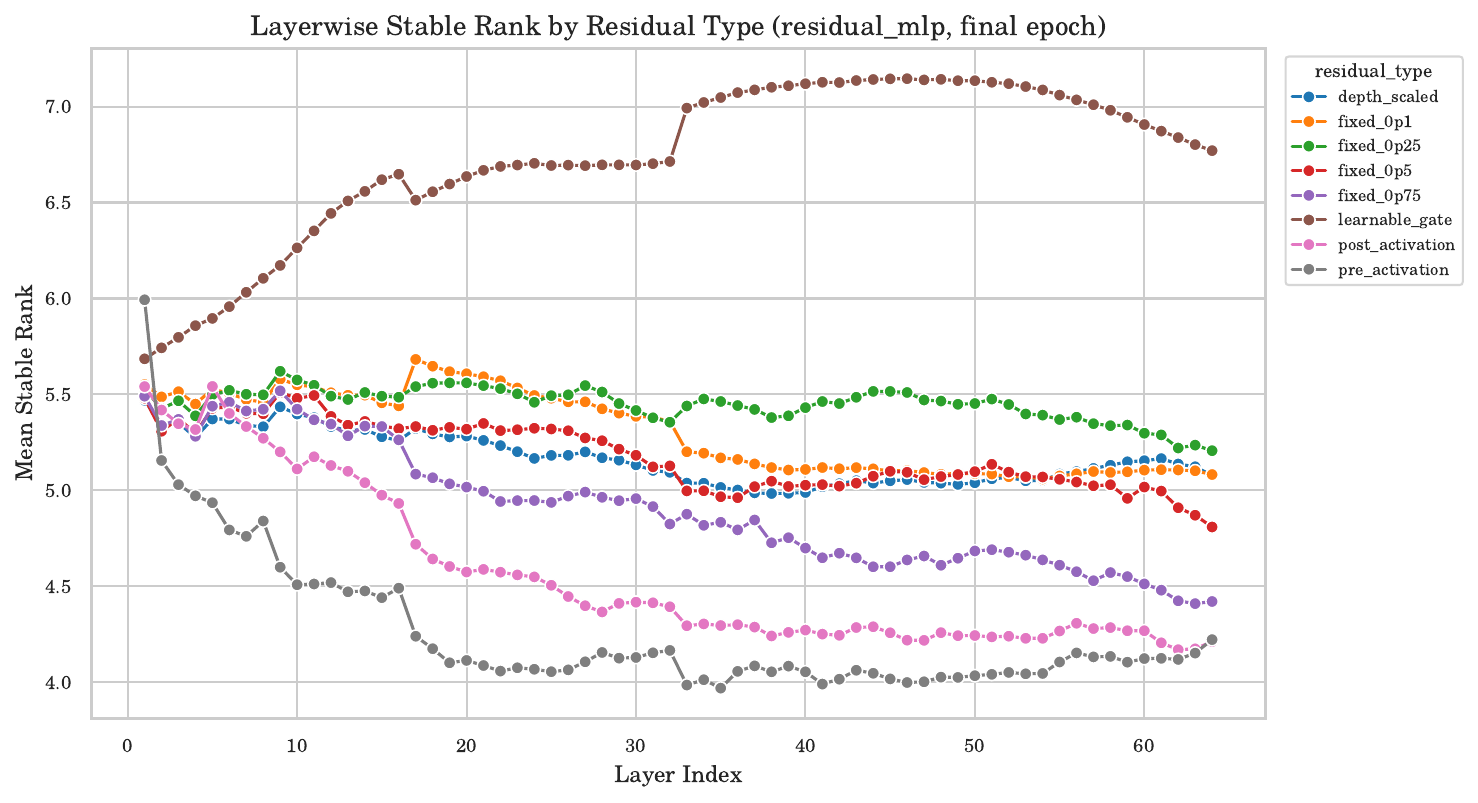}
\captionsetup{font=footnotesize,skip=2pt}
\captionof{figure}{Observation on Layer-wise Stable rank categorized by different residual implementation styles of Residual MLP network type. Values are calculated at the final epoch so as to find which implementation style would preserve rank to the deepest layer. Values are the mean over Residual-MLP runs that have a layer of the given index.}
\label{fig:residual_type_vs_depth_stable_rank}

\vspace{2mm}

\centering
\includegraphics[width=1.2\columnwidth,height=1.5\textheight,
keepaspectratio]{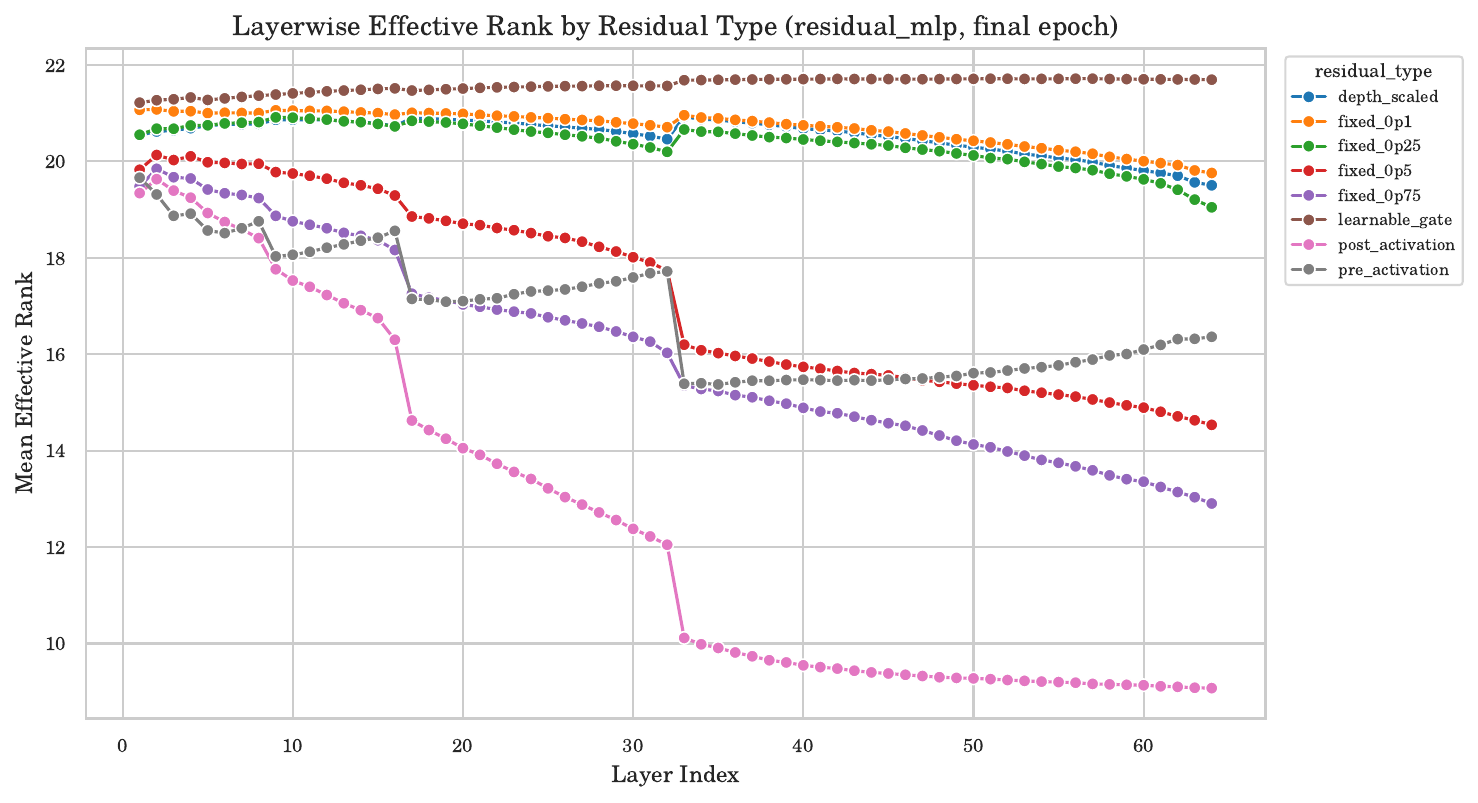}
\captionsetup{font=footnotesize,skip=2pt}
\captionof{figure}{Observation on Layer-wise Effective rank categorized by different residual implementation styles of Residual MLP network type. Values are calculated at the final epoch so as to find which implementation style would preserve rank to the deepest layer. Values are the mean over Residual-MLP runs that have a layer of the given index.}
\label{fig:residual_type_vs_depth_effective_rank}

\vspace{2mm}

\section{Choice of the DRC Thresholds}
\label{sec: support_analysis_DRC}
\paragraph{Justification of Threshold Parameters.} The exact quantitative bounds defining DRC ($\text{SR}_{\min} \le 1.05$, $\text{SR}_{\text{final}} \le 1.05$, $\text{ER}_{\text{final}} \le 1.5$) are grounded in both singular value spectral theory and empirical performance phase transitions across $N = 4,390$ models:
\begin{enumerate}
    \item \textbf{Stable Rank Threshold ($\text{SR} \leq 1.05$)}: SR $=1$ exactly when the representation is a single direction ($\sigma_2 = 0$). SR $\leq 1.05$ means the leading singular value carries at least $1/1.05 \approx 95.24\%$ of the squared Frobenius mass. We use this margin as an empirical tolerance for small residual spectral mass; it is not a floating-point tolerance. Requiring both $\text{SR}_{\min} \leq 1.05$ and $\text{SR}_{\text{final}} \leq 1.05$ excludes runs that recover.
    
    \item \textbf{Effective Rank Threshold ($\text{ER} \le 1.5$)}: Effective Rank is defined as the exponentiated Shannon entropy of the normalized singular values, where $p_i=\sigma_i/\sum_k \sigma_k$:
\begin{equation}
\operatorname{ER}(A)=\exp\left(-\sum_{i=1}^{R} p_i\ln p_i\right).
\end{equation}
\end{enumerate}

An upper bound of $ER \leq 1.50$ corresponds to a maximum spectral entropy of
$H(p) \leq \ln(1.50) \approx 0.4055$ nats. At this entropy level, the
leading singular value $\sigma_1$ accounts for more than $85\%$-$95\%$ of
the total singular value mass, leaving only a small fraction distributed
across the remaining singular values. Consequently, the feature
representation becomes effectively confined to a narrow one-dimensional
linear corridor, irrespective of the ambient feature dimension.

\paragraph{Emperical Check}
A rank-one final representation cannot separate more than one class, making chance-level accuracy inevitable for any $k>1$ classification task. The 192-run CIFAR-10 Plain CNN sweep illustrates this: all runs with $ER_{\mathrm{final}} \leq 1.5$ remain at chance accuracy ($\leq 10.15\%$), while runs with $ER_{\mathrm{final}}>1.5$ reach up to $91.01\%$ test accuracy. Since the threshold was selected using this same sweep, these results serve as a consistency check rather than an independent validation.

\centering
\includegraphics[width=\columnwidth]{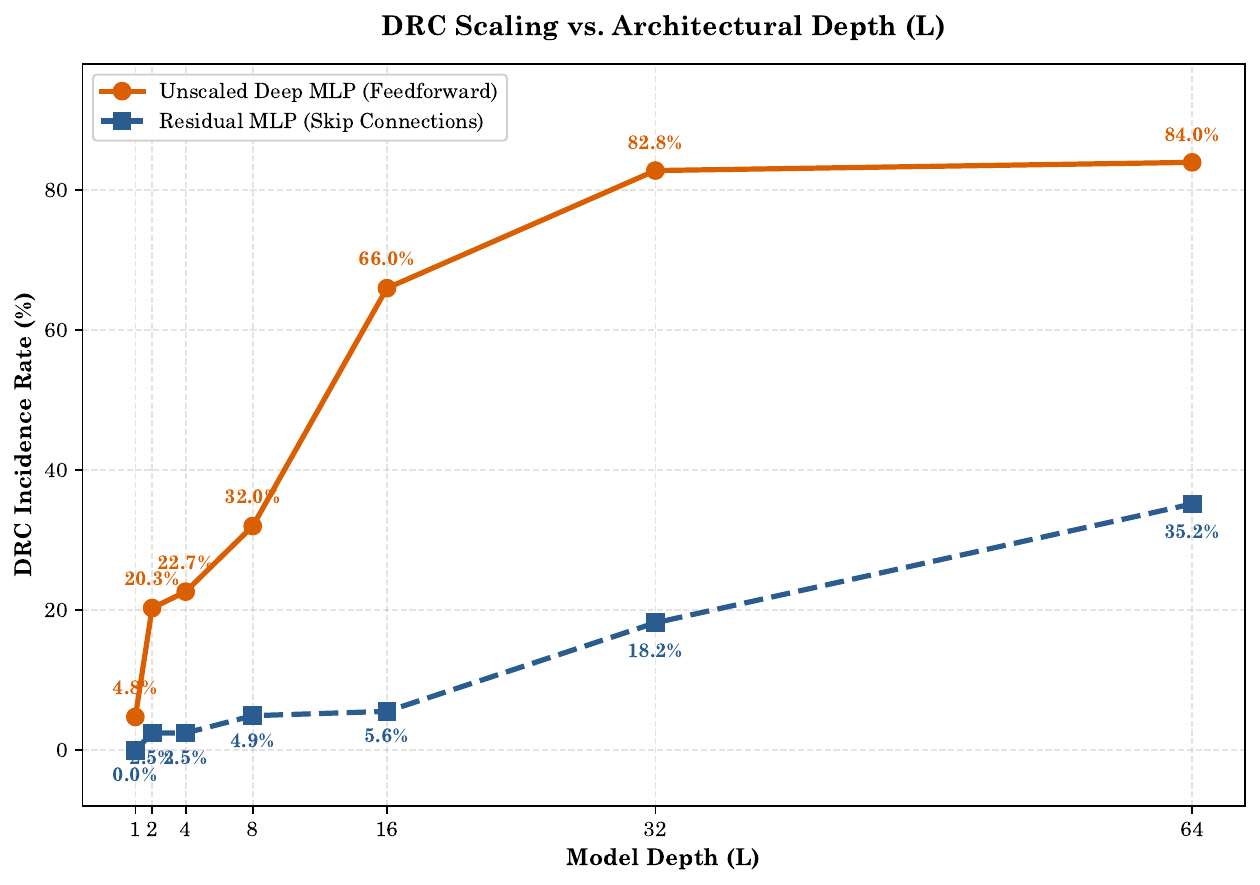}
\captionof{figure}{
\textbf{DRC incidence versus model depth.}
DRC incidence (\%) increases nearly monotonically with model depth in both unscaled Deep Plain MLPs (without depth-dependent scaling such as $1/\sqrt{L}$ or ReZero gating) and Residual MLPs.
}
\label{fig:drc_incidence_rate_depth}

\vspace{2mm}

\centering
\includegraphics[width=\columnwidth]{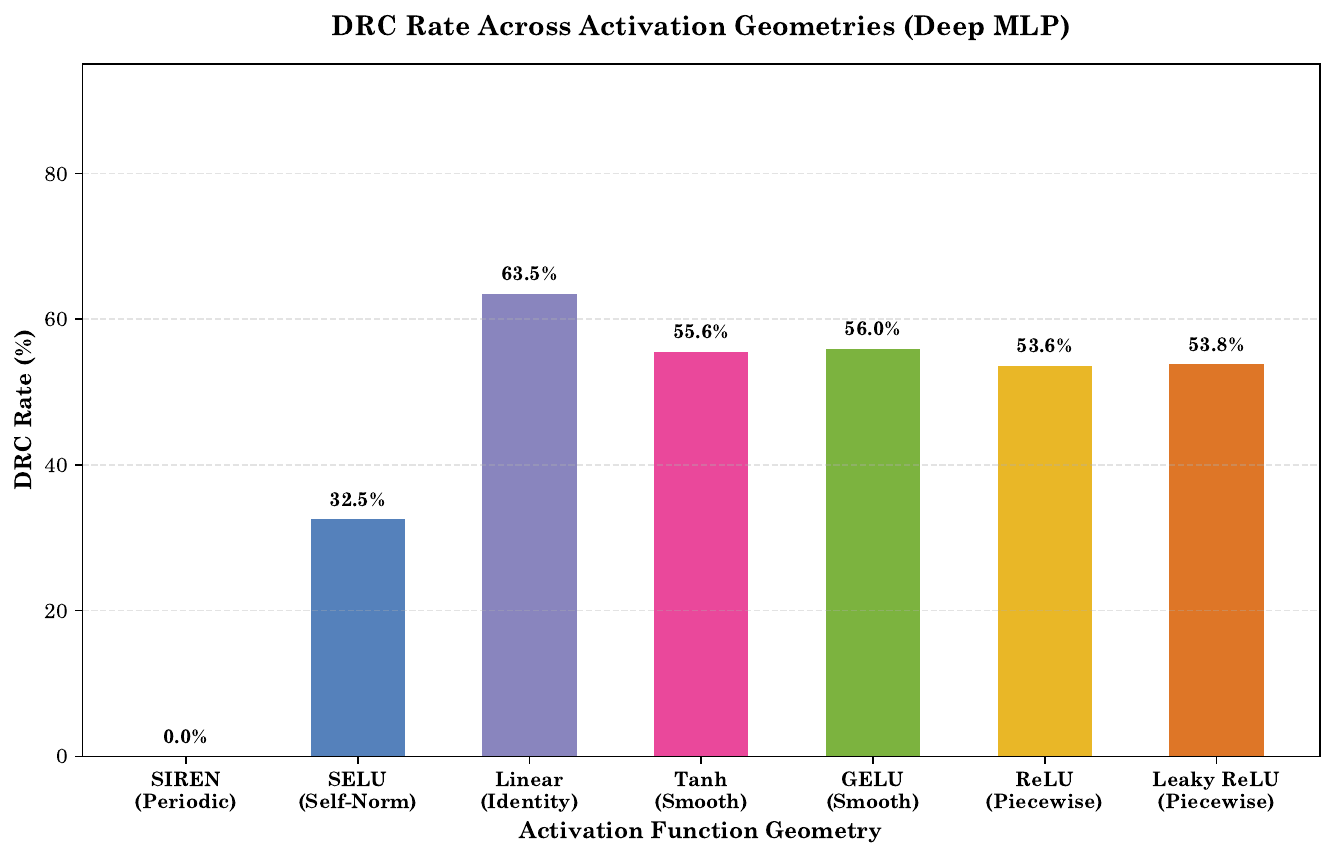}
\captionof{figure}{
\textbf{Activation-wise susceptibility to DRC.}
DRC incidence across activation functions for $N_{\mathrm{models}}=1,764$ models. SIREN shows $0\% $DRC yet has the lowest success (Table~\ref{tab:factors_multivariate})
}
\label{fig:drc_activation_vulnerability}

\vspace{2mm}

\begin{center}
\captionof{table}{\textbf{Milestone statistics for a benign low-rank trajectory.} The network maintains low stable rank ($\mathrm{SR}\approx1.05$--1.19) while preserving effective rank ($\mathrm{ER}\ge10.4$), enabling it to achieve 90.35\% CIFAR-10 test accuracy.}
\label{tab:benign_low_rank_milestones}

\fontsize{6.5}{7.5}\selectfont
\setlength{\tabcolsep}{3pt}

\resizebox{0.98\columnwidth}{!}{
\begin{tabular}{ccccccc}
\toprule
\textbf{Epoch} &
\textbf{Acc. (\%)} &
\textbf{Min SR} &
\textbf{Mean SR} &
\textbf{Min ER} &
\textbf{Mean ER} &
\textbf{Loss} \\
\midrule
1   & 9.92  & 1.191 & 2.245 & 13.54 & 165.85 & 2.515 \\
5   & 12.45 & 1.102 & 1.813 & 13.97 & 143.40 & 2.336 \\
10  & 47.62 & 1.213 & 1.865 & 14.19 & 88.25  & 1.429 \\
25  & 76.97 & 1.163 & 2.136 & 18.50 & 64.85  & 0.705 \\
50  & 83.49 & 1.129 & 2.654 & 19.77 & 53.43  & 0.520 \\
100 & 86.54 & 1.109 & 2.637 & 18.27 & 44.83  & 0.511 \\
150 & 89.82 & 1.082 & 2.749 & 16.16 & 40.25  & 0.386 \\
200 & 87.47 & 1.072 & 3.201 & 13.95 & 37.98  & 0.439 \\
250 & 90.63 & 1.058 & 3.138 & 11.61 & 35.41  & 0.350 \\
300 & 90.35 & 1.051 & 3.040 & 10.39 & 33.11  & 0.362 \\
\bottomrule
\end{tabular}
}
\end{center}

\centering
\includegraphics[width=\columnwidth]{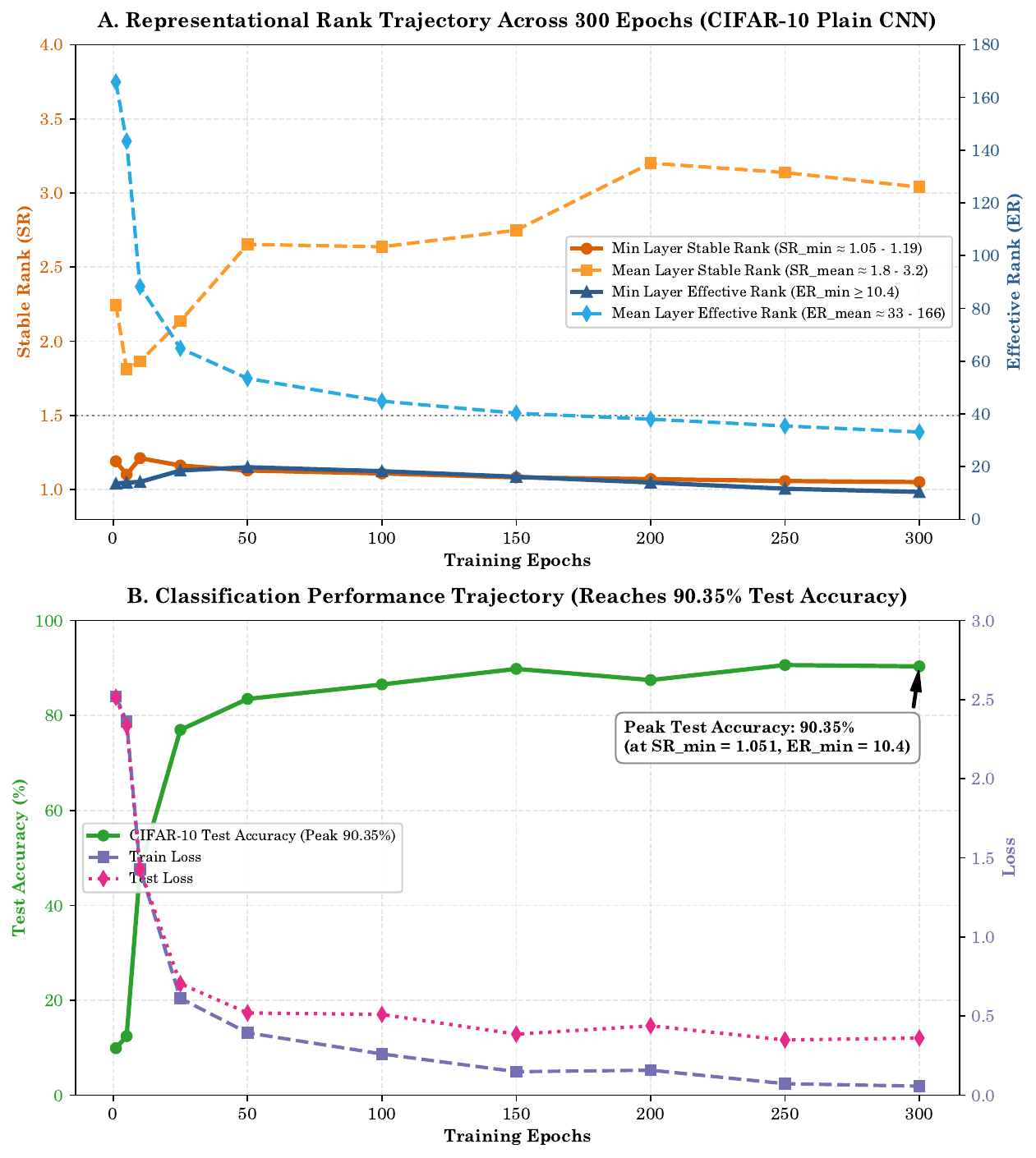}
\captionof{figure}{
\textbf{Low stable rank does not imply DRC.}
Despite $\mathrm{SR}_{\min}\approx1.05$, preserved effective rank ($\mathrm{ER}_{\min}\ge10.4$) enables the model to reach 90.35\% CIFAR-10 test accuracy.
}
\label{fig:lowstable_higheffective_goodperformance_Drc}

\vspace{2mm}

\centering
\includegraphics[width=\columnwidth]{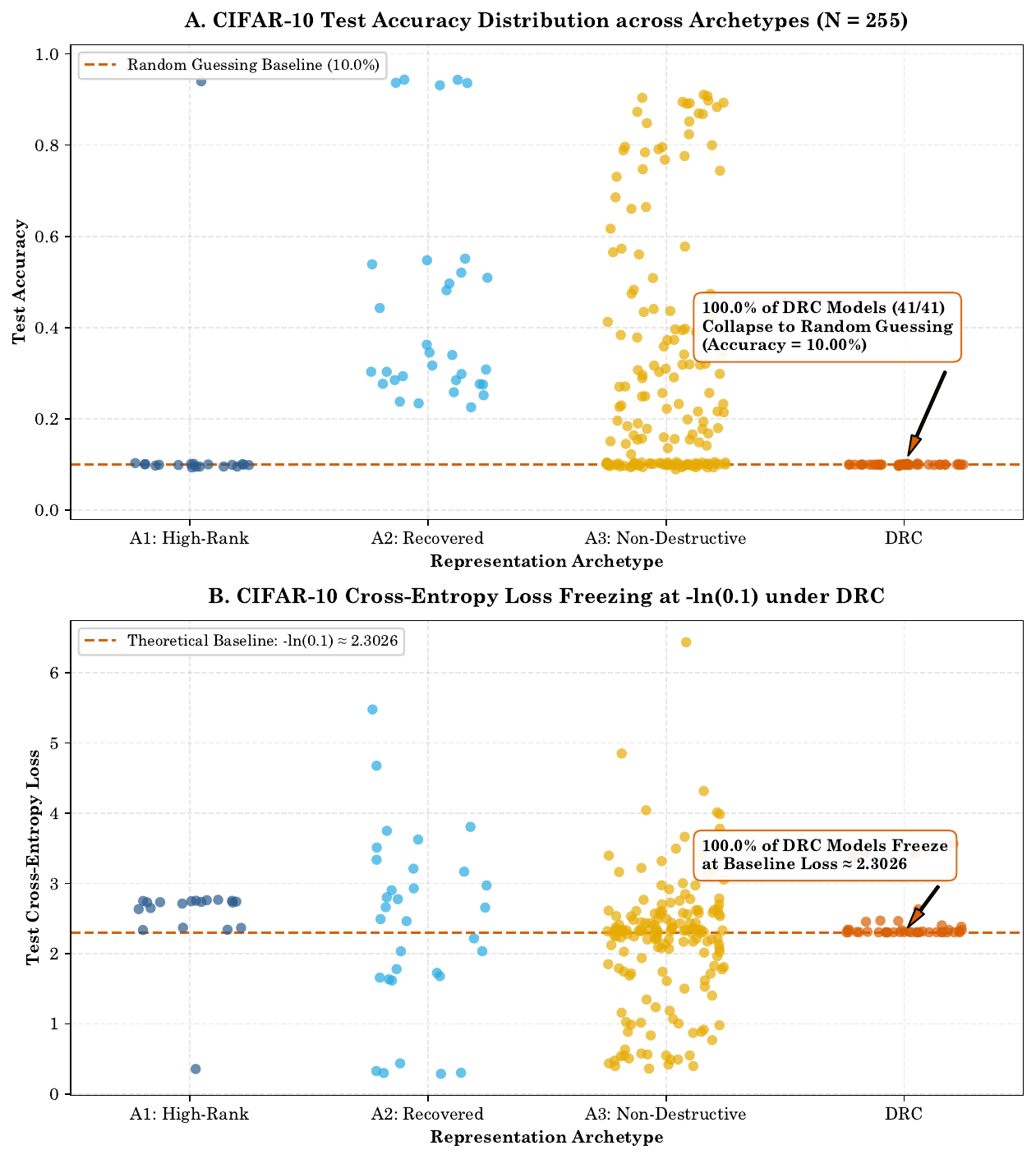}
\captionof{figure}{
\textbf{Classification performance across A1--A3 and DRC.}
A1--A3 achieve high test accuracy, whereas all DRC models remain at random-guessing accuracy ($\approx10\%$) with cross-entropy fixed at the theoretical baseline ($-\ln(0.1)\approx2.3026$).
ViT runs are analyzed separately in Appendix~\ref{app:extension to Vision Transformers}} 
\label{fig:cifar10_random_guessing}

\vspace{2mm}

\begin{table}[h]
\centering
\caption{\textbf{Common hyperparameter specifications across synthetic and vision experiment sweeps.}}
\label{tab:common_hyperparameters}
\fontsize{6.5}{7.5}\selectfont
\setlength{\tabcolsep}{3pt}
\renewcommand{\arraystretch}{0.95}
\resizebox{0.98\columnwidth}{!}{
\begin{tabular}{p{0.28\columnwidth} p{0.34\columnwidth} p{0.30\columnwidth}}
\toprule
\textbf{Hyperparameter} & \textbf{Synthetic Sweeps ($N=4{,}084$)} & \textbf{CIFAR-10 Plain CNN sweep ($N=192$)} \\
\midrule
Random Seed & 42 & 42 \\
Learning Rate ($\eta$) & 0.001 & 0.001 \\
Weight Decay ($\lambda$) & 0.0 & 0.0001 \\
Batch Size ($B$) & Full-batch (32) & 128 \\
Optimizers & Adam, Momentum SGD, SGD & Adam, SGD \\
Loss Criterion & Mean Squared Error (MSE) & Cross-Entropy Loss \\
Input/Output Dimension & $64\rightarrow64$ & $3\times32\times32\rightarrow10$ \\
\bottomrule
\end{tabular}
}
\end{table}

\section{Statistical Background}
\label{sec: statistical background}
This section briefly summarizes the statistical measures used throughout the paper.

\subsection{Odds Ratio (OR)}

The \emph{Odds Ratio (OR)} quantifies the strength of association between two groups. It is defined as the ratio of their odds of success:

\begin{equation}
\mathrm{OR}
=
\frac{\dfrac{p_A}{1-p_A}}
{\dfrac{p_B}{1-p_B}},
\end{equation}

where $p_A$ and $p_B$ denote the probabilities of success for Groups A and B, respectively.

Interpretation:
\begin{itemize}
    \item $\mathrm{OR}=1$: no association.
    \item $\mathrm{OR}>1$: Group A has higher odds of success.
    \item $\mathrm{OR}<1$: Group A has lower odds of success.
\end{itemize}

\subsection{Adjusted Odds Ratio (AOR)}

When multiple variables may jointly influence the outcome, a multivariate logistic regression model is used:

\begin{equation}
\log\!\left(\frac{p}{1-p}\right)
=
\beta_0+\beta_1X_1+\beta_2X_2+\cdots+\beta_kX_k.
\end{equation}

The adjusted odds ratio for predictor $X_i$ is

\begin{equation}
\mathrm{AOR}_i=\exp(\beta_i),
\end{equation}

which represents the independent effect of $X_i$ while controlling for all other predictors.

\subsection{Fisher's Exact Test}

Fisher's Exact Test evaluates the association between two categorical variables in a $2\times2$ contingency table. Unlike the $\chi^2$ test, it computes the exact probability of observing the contingency table under the null hypothesis:

\begin{equation}
\text{P(table)}=
\frac{(a+b)!(c+d)!(a+c)!(b+d)!}
{a!\,b!\,c!\,d!\,n!},
\end{equation}

where $n=a+b+c+d$.

The two-sided p-value sums P over all tables with the same margins that are no more probable than the observed table

\subsection{Confidence Intervals and \texorpdfstring{$p$}--values}

A $95\%$ confidence interval (CI) provides a plausible range for the true population effect size. For odds ratios, statistical significance is typically inferred when the $95\%$ CI does not include 1. The confidence interval is computed as

\begin{equation}
\exp(\beta_i \pm 1.96\,\mathrm{SE}(\beta_i)).
\end{equation}

The $p$-value denotes the probability of observing an effect at least as extreme as the measured one under the null hypothesis. Conventionally,

\begin{itemize}
    \item $p<0.05$: statistically significant,
    \item $p<0.001$: highly statistically significant.
\end{itemize}

\subsection{Receiver Operating Characteristic (ROC) Curve and AUC}

A Receiver Operating Characteristic (ROC) curve plots the True Positive Rate (Sensitivity) against the False Positive Rate (1--Specificity) across all classification thresholds.

The Area Under the ROC Curve (AUC) summarizes overall classification performance:

\begin{itemize}
    \item $\mathrm{AUC}=0.5$: random prediction.
    \item $\mathrm{AUC}=1.0$: perfect prediction.
\end{itemize}

\section{Early-Trajectory Prediction of Optimization Outcomes}
\label{app:early_trajectory_prediction_of_optimization_outcomes}

To examine whether representation dynamics contain information about optimization outcomes before convergence, we classified models using trajectory information available at early training milestones of $10\%$, $25\%$, and $50\%$ of training. For synthetic MLPs, these correspond to epochs 100, 250, and 500 of a 1,000-epoch run. Models were assigned to an early trajectory category based on the minimum stable rank observed up to each milestone and its subsequent change: \textbf{Never Collapsed} if $SR_t > 1.50$ throughout the observed interval, \textbf{Early Recovery} if the representation had collapsed but subsequently expanded, and \textbf{Early Stuck} if it remained near the collapse bottleneck. The classification uses only information available up to the respective milestone and does not use final loss or final representations.

Across $2,268$ synthetic MLP runs, the early trajectory categories differ in their eventual optimization outcomes. At $25\%$ of training, 618 models were classified as Early Recovery, with a $35.8\%$ success rate, compared with $4.9\%$ among the $930$ Early Stuck models. This corresponds to a relative risk of $7.16\times$ and an odds ratio of $10.60$. Similar differences are observed at $10\%$ and $50\%$, with odds ratios of $6.33$ and $12.11$, respectively. These comparisons are unadjusted for configuration (activation, optimizer, depth, initialization) and should be read as descriptive.

Overall, these results suggest that representation trajectories contain information that is related to eventual optimization outcomes before convergence. In particular, models that remain near the rank-collapse bottleneck early in training tend to have lower subsequent success rates, whereas models showing early recovery tend to have better outcomes. This result is consistent with early trajectory carrying forecasting information, we do not yet test whether it can be turned into a reliable automated early-stopping criterion, which we leave to future work.

\subsection{Cross Validated association}
\label{subsec: cross_validated_association}
\begin{table}[H]
\centering
\caption{\textbf{5-fold stratified cross-validation performance ($M_\mathrm{models}=4{,}084$).} SD=Standard deviation}
\label{tab:cv_classifier_performance}

\fontsize{6.5}{7.5}\selectfont
\setlength{\tabcolsep}{2.5pt}
\renewcommand{\arraystretch}{0.92}

\resizebox{0.98\columnwidth}{!}{
\begin{tabular}{lcccc}
\toprule
\textbf{Model} &
\textbf{Mean AUC} &
\textbf{SD} &
\textbf{95\% CI} &
\textbf{Acc.} \\
\midrule
\textsuperscript{1}Static Final-Rank Endpoint & 0.7743 & 0.0193 & [0.7574, 0.7912] & 69.20\% \\
\textsuperscript{2}Hyperparameter Controls Only & 0.9459 & 0.0073 & [0.9395, 0.9523] & 90.77\% \\
\textbf{\textsuperscript{3}Full Trajectory + Controls} & \textbf{0.9660} & \textbf{0.0048} & \textbf{[0.9618, 0.9702]} & \textbf{92.58\%} \\
\bottomrule
\end{tabular}
}
\vspace{1mm}
{\fontsize{6}{7}\selectfont
\textsuperscript{1}Uses only final-epoch rank snapshots ($\text{SR}_{\text{final}}, \text{ER}_{\text{final}}$).  \textsuperscript{2}Uses setup features only (activation, architecture, depth, optimizer, rank).  \textsuperscript{3}Combines temporal trajectory archetypes with setup features.
}

\end{table}

\begin{figure}[ht]
    \centering
    \includegraphics[
        width=0.8\textwidth,
        keepaspectratio
    ]{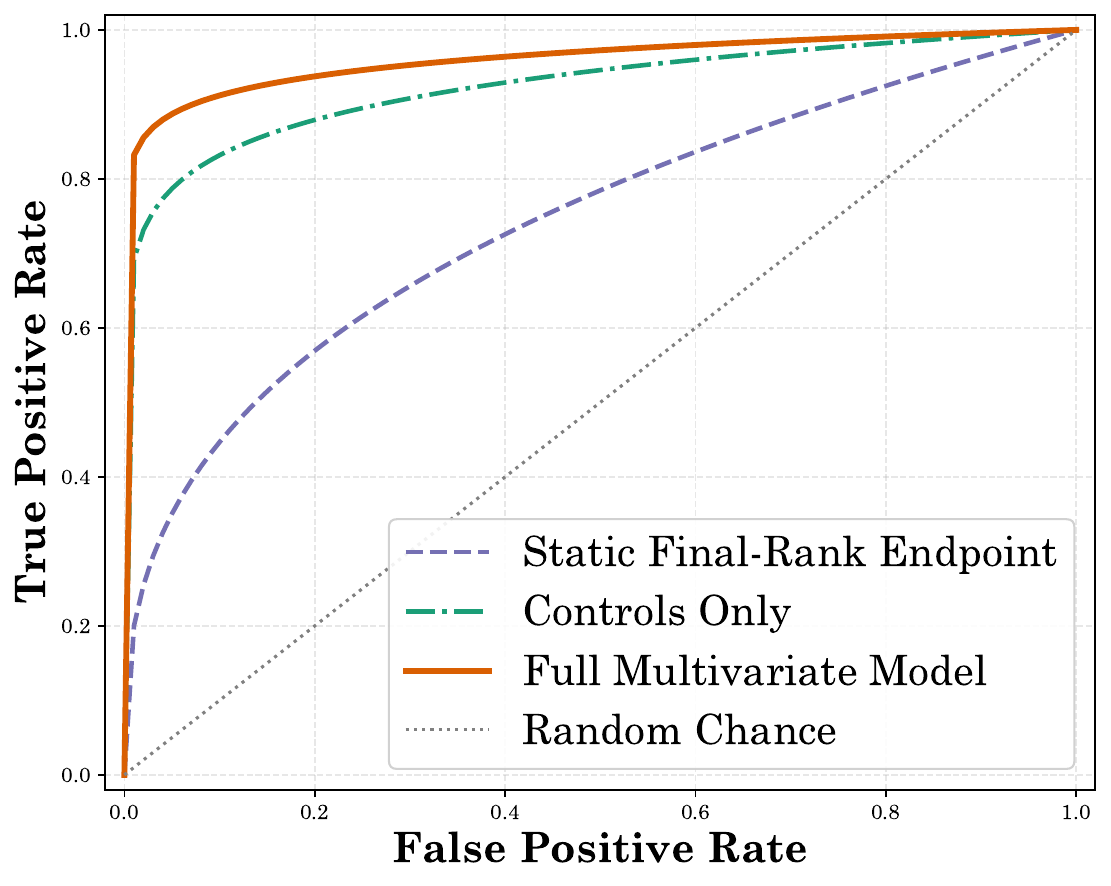}

    \caption{5-Fold Cross-Validation ROC Curve Comparison}
    \label{fig:classifier_roc_cv}
\end{figure}

\section{Sensitivity to Trajectory Classification Thresholds}
\label{app:sensitivity_to_trajectory_classification_thresholds}
This is a robustness check of the early-milestone forecasting result in Appendix~\ref{app:early_trajectory_prediction_of_optimization_outcomes}.
To examine how sensitive the trajectory–outcome relationship is to the chosen numerical boundaries, we repeated the archetype classification at the $25\%$ training milestone $(epoch 250)$ across a grid of collapse and recovery thresholds. The collapse threshold $\tau_{\mathrm{collapse}}$ was varied over $\{1.05,1.10,1.15,1.20,1.50\}$, and the recovery threshold $\Delta\tau_{\mathrm{rec}}$ over $\{0.05,0.10,0.15,0.20\}$, giving $20$ threshold combinations.

Across this grid, the relationship between trajectory type and optimization outcome is similar across the tested settings. Recovery trajectories have higher success rates than early-stuck trajectories, with odds ratios ranging from $7.35$ to $15.71$ and relative risks from $5.25$ to $10.69$. Recovery success rates range from $31.4\%$ to $37.7\%$, compared with $3.1\%$ to $6.8\%$ for the early-stuck group. Changing the thresholds therefore changes the number of models assigned to each archetype, but the separation between recovery and persistent low-rank trajectories remains present across the tested boundaries.

\section{Extension to V ision Transformers}
\label{app:extension to Vision Transformers}
To examine whether the representation trajectory framework also extends beyond feedforward and convolutional architectures, we evaluated 48 CIFAR-10 Vision Transformer runs using both standard residual and no-residual configurations at depths $L \in \{3,6,9\}$. We tracked Stable Rank (SR) and Effective Rank (ER) throughout training and classified the early trajectories using only the information available up to the corresponding training milestone.

\begin{table}[t]
\centering
\caption{Depth-Wise Representation \& Performance Scaling ($N=48$ Models, 200 Epochs)}
\label{tab:depth-wise-scaling}
\resizebox{\columnwidth}{!}{%
\begin{tabular}{lcccccccc}
\hline
\textbf{Architecture} & \textbf{Depth ($L$)} & \textbf{$N$} & \textbf{Train Acc (\%)} & \textbf{Test Acc (\%)} & \textbf{Deepest Block SR} & \textbf{Deepest Block ER} & \textbf{Penult SR} & \textbf{Catastrophic DRC} \\
\hline
\textbf{Standard ViT} \textit{(Pre-LN + Residual)}
& 3 & 8 & $87.0\pm8.7$ & $76.2\pm3.6$ & $6.50\pm1.85$ & $72.7\pm23.8$ & $5.19\pm0.49$ & \textbf{0 / 8 (0.0\%)} \\

\textbf{Standard ViT} \textit{(Pre-LN + Residual)}
& 6 & 8 & $83.2\pm17.5$ & $72.3\pm10.6$ & $6.27\pm2.00$ & $71.2\pm22.8$ & $5.08\pm1.09$ & \textbf{0 / 8 (0.0\%)} \\

\textbf{Standard ViT} \textit{(Pre-LN + Residual)}
& 9 & 8 & $91.9\pm8.6$ & $77.2\pm3.6$ & $7.25\pm1.35$ & $75.1\pm25.4$ & $5.79\pm0.78$ & \textbf{0 / 8 (0.0\%)} \\

\textbf{No-Residual ViT} \textit{(Pre-LN Only)}
& 3 & 8 & $74.3\pm6.8$ & $68.0\pm4.8$ & $3.53\pm0.51$ & $29.7\pm26.5$ & $3.58\pm0.59$ & \textbf{0 / 8 (0.0\%)} \\

\textbf{No-Residual ViT} \textit{(Pre-LN Only)}
& 6 & 8 & $66.6\pm8.8$ & $62.8\pm7.8$ & $2.94\pm0.62$ & $22.9\pm19.7$ & $3.29\pm0.27$ & \textbf{0 / 8 (0.0\%)} \\

\textbf{No-Residual ViT} \textit{(Pre-LN Only)}
& 9 & 8 & $52.8\pm27.0$ & $49.2\pm24.6$ & $2.46\pm1.04$ & $13.5\pm11.6$ & $2.62\pm1.07$ & \textbf{2 / 8 (25.0\%)} \\
\hline
\end{tabular}%
}
\end{table}

\paragraph{}Standard residual ViTs maintain substantially higher representation rank across depth, while no-residual ViTs show a progressive contraction in rank. At depth 9, the no-residual models have substantially lower effective rank than their residual counterparts, with 2 of the 8 evaluated runs satisfying the DRC criterion. Both of these runs reach approximately rank-one representations and achieve chance-level accuracy on CIFAR-10. See Table~\ref{tab:depth-wise-scaling}. \\

\begin{table*}[t]
\centering
\caption{Layer-by-Layer Metrics across Depth at $L=9$ ($N=16$ Runs, Epoch 200)}
\label{tab:layer-wise-metrics}
\resizebox{\textwidth}{!}{%
\begin{tabular}{lcccc}
\hline
\textbf{Layer} & \textbf{Standard ViT SR} & \textbf{Standard ViT ER} & \textbf{No-Residual ViT SR} & \textbf{No-Residual ViT ER} \\
\hline
\textbf{Block 1} & $5.95\pm1.19$ & $57.7\pm13.8$ & $3.49\pm2.39$ & $39.7\pm27.1$ \\
\textbf{Block 2} & $7.08\pm1.40$ & $68.8\pm20.2$ & $3.00\pm2.35$ & $29.4\pm31.3$ \\
\textbf{Block 3} & $7.40\pm1.92$ & $73.3\pm22.8$ & $2.66\pm1.90$ & $21.6\pm26.8$ \\
\textbf{Block 4} & $8.10\pm1.68$ & $75.4\pm23.4$ & $2.55\pm1.65$ & $15.7\pm18.8$ \\
\textbf{Block 5} & $8.34\pm1.94$ & $76.6\pm23.8$ & $2.68\pm1.75$ & $12.4\pm13.9$ \\
\textbf{Block 6} & $8.82\pm2.43$ & $77.2\pm23.9$ & $2.78\pm1.81$ & $11.5\pm12.2$ \\
\textbf{Block 7} & $8.54\pm1.99$ & $76.9\pm23.2$ & $2.65\pm1.70$ & $11.3\pm11.9$ \\
\textbf{Block 8} & $8.69\pm1.67$ & $76.8\pm23.3$ & $2.67\pm1.75$ & $11.7\pm11.9$ \\
\textbf{Block 9} \textit{(Deepest)} & $7.25\pm1.27$ & $75.1\pm23.7$ & $2.46\pm0.97$ & $13.5\pm10.9$ \\
\textbf{Penultimate Layer} & $5.79\pm0.73$ & $59.2\pm20.4$ & $2.62\pm1.00$ & $13.4\pm10.7$ \\
\hline
\end{tabular}%
}
\end{table*}

\paragraph{}The layer-wise analysis further shows that this difference builds up with depth. In the evaluated depth-9 models, the effective rank of the no-residual configuration decreases from approximately 39.7 at the first block to 11.3 at the most contracted layer, while the corresponding residual models maintain substantially higher effective rank. This provides an empirical example of depth-dependent representation contraction occurring when skip connections are absent. See Table~\ref{tab:layer-wise-metrics}. \\

\begin{table*}[t]
\centering
\caption{Early Trajectory Archetype Forecasting in Vision Transformers}
\label{tab:early-trajectory-archetype}
\resizebox{\textwidth}{!}{%
\begin{tabular}{lccccc}
\hline
\textbf{Training Milestone} & \textbf{Arch 1: Never Collapsed (Success \%)} & \textbf{Arch 2: Early Recovery (Success \%)} & \textbf{Arch 3: Early Stuck (Success \%)} & \textbf{Odds Ratio (Arch 2 vs 3)} & \textbf{Mean Final Accuracy (A2 vs A3)} \\
\hline
\textbf{10\% of Training} \textit{(Epoch 20)}
& 31 ($90.3\%$) & \textbf{14 ($78.6\%$)} & 3 ($0.0\%$) & \textbf{$23.00\times$} & $64.4\%$ vs $25.9\%$ \\

\textbf{25\% of Training} \textit{(Epoch 50)}
& 31 ($90.3\%$) & \textbf{15 ($73.3\%$)} & 2 ($0.0\%$) & \textbf{$12.78\times$} & \textbf{$64.0\%$ vs $10.0\%$} \\

\textbf{50\% of Training} \textit{(Epoch 100)}
& 31 ($90.3\%$) & \textbf{15 ($73.3\%$)} & 2 ($0.0\%$) & \textbf{$12.78\times$} & \textbf{$64.0\%$ vs $10.0\%$} \\
\hline
\end{tabular}%
}
\end{table*}

\paragraph{}This replicates, in an independent architecture family, the early-forecasting direction observed in the synthetic MLP sweeps (Appendix~\ref{app:early_trajectory_prediction_of_optimization_outcomes}): at $25\%$ of training, 11 of 15 Early Recovery models eventually succeed, compared with 0 of 2 Early Stuck/DRC models. The stuck-group count is small ($n=2$), so the magnitude should be treated as indicative rather than precise; the consistency in direction with the synthetic results is the main takeaway. See Table~\ref{tab:early-trajectory-archetype}.

\section{Dataset Configuration}
\label{sec:data configuration}
\subsection{Synthetic Low-Rank Regression:}
The data description in detail has been already mentioned in the main section of the paper. 
\subsection{CIFAR- 10 Image Classification:}
\textbf{Dataset:}$50,000$ training and $10,000$ testing images across $10$ classes are used.
\textbf{Preprocessing:}Normalization using CIFAR-10 means and standard deviations. Training images are augmented with random crops and random horizontal flips.

\section{Experimental Sweeps Census and Architectural Specifications}
\label{sec:architectural specifications}
\subsection{Architectural Specifications}
Optimization algorithms used - Table~\ref{tab:optimizers}\\
Initialization Schemes used - Table~\ref{tab:initialization}\\
Activation Functions used - Table~\ref{tab:activations}\\
Residual Connection Variants used - Table~\ref{tab:residual_types}\\
Description of Plain MLP and DeepLinear MLP architectures used - Table~\ref{tab:plainmlp_deeplinearmlp}\\
Description of Residual MLP used. - Table~\ref{tab:residualmlpdesc}\\
Description of Plain CNN used. - Table~\ref{tab:plainCNN_desc}\\
Description of ResNet-18 and ResNet-50 used - Table~\ref{tab:resnet1850_desc}\\
Description of ViT used - Table~\ref{tab:vit_description}

\begin{table}[H]
\captionof{table}{Optimization algorithms used in the experimental study. All optimizers use learning rate $\eta=10^{-3}$ and weight decay $\lambda=0$ (synthetic), $10^{-4}$ (CIFAR-10 CNN)}
\label{tab:optimizers}
\small
\begin{tabular}{p{0.18\columnwidth}p{0.28\columnwidth}p{0.42\columnwidth}}
\toprule
\textbf{Optimizer} &
\textbf{Description} &
\textbf{Default Hyperparameters} \\
\midrule
SGD &
Plain gradient descent &
--- \\

SGD + Momentum &
Momentum-based GD &
$\beta=0.9$ \\

Adam &
Adaptive moment estimation &
$\beta_1=0.9$, $\beta_2=0.999$, $\epsilon=10^{-8}$ \\
\bottomrule
\end{tabular}
\end{table}

\begin{table}[H]
\centering
\captionof{table}{Initialization schemes evaluated.}
\label{tab:initialization}
\small
\begin{tabular}{lll}
\toprule
\textbf{Scheme} & \textbf{Initialization} & \textbf{Key Parameter} \\
\midrule
Default &
PyTorch Kaiming Uniform &
-- \\
Low-Rank &
$W=\dfrac{AB}{\sqrt{rd_{\mathrm{in}}}},\;
A,B\sim\mathcal{N}(0,1)$ &
Rank $r$ ($r=1$) \\
\bottomrule
\end{tabular}
 \end{table}

\begin{table}[H]
\begin{center}
\captionof{table}{Activation functions used in the experiments.}

\label{tab:activations}
\small
\begin{tabular}{ll}
\toprule
\textbf{Activation} & \textbf{Definition} \\
\midrule
Linear & $\Phi(x)=x$ \\
ReLU & $\Phi(x)=\max(0,x)$ \\
Leaky ReLU & $\Phi(x)=\max(0.01x,x)$ \\
Tanh & $\Phi(x)=\tanh(x)$ \\
GELU & Standard GELU approximation \\
SELU & Standard SELU activation \\
SIREN & $\Phi(x)=\sin(\omega_0x),\ \omega_0=30$ \\
\bottomrule
\end{tabular}
\end{center}
\end{table} 

\begin{table}[H]
\begin{center}
\captionof{table}{Residual connection variants. Depth scaled variant defined for $L \geq 2$.}
\label{tab:residual_types}
\small
\begin{tabular}{ll}
\toprule
\textbf{Variant} & \textbf{Residual Mapping} \\
\midrule
Post-activation & $h^{(l)}=h^{(l-1)}+\Phi(W^{(l)}h^{(l-1)})$ \\
Pre-activation & $h^{(l)}=h^{(l-1)}+W^{(l)}\Phi(h^{(l-1)})$ \\
Depth-scaled & $h^{(l)}=h^{(l-1)}+\frac{1}{L-1}\Phi(\cdot)$ \\
Fixed-scale & $h^{(l)}=h^{(l-1)}+\alpha\Phi(\cdot)$ \\
Learnable gate (ReZero-style) & $h^{(l)}=h^{(l-1)}+\alpha_l\Phi(\cdot)$ \\
\bottomrule
\end{tabular}
\end{center}
\end{table}

\begin{table}[H]
\begin{center}
\captionof{table}{Description of Plain MLP and DeepLinear MLP architectures used}
\label{tab:plainmlp_deeplinearmlp}
\footnotesize
\begin{tabular}{c}
\toprule
\textbf{Plain MLP / DeepLinear} \\
\midrule
Input: Flattened Vector \\
Linear ($d_{\mathrm{in}}\rightarrow H$) \\
$(L-1)\times$ Linear ($H\rightarrow H$) \\
Linear ($H\rightarrow d_{\mathrm{out}}$) \\
\midrule
DeepLinear uses identity activation throughout. \\
\textit{Note: $L$ counts hidden layers, the network has $L+1$ weight matrices.} \\
\bottomrule
\end{tabular}
\end{center}
\end{table}

\begin{table}
\begin{center}
\captionof{table}{Description of Residual MLP used.}
\label{tab:residualmlpdesc}
\footnotesize
\begin{tabular}{c}
\toprule
\textbf{ResidualMLP} \\
\midrule
Input: Flattened Vector \\
Linear ($d_{\mathrm{in}}\rightarrow H$) \\
$(L-1)\times$ Residual Block ($H\rightarrow H$) \\
Linear ($H\rightarrow d_{\mathrm{out}}$) \\
\midrule
Residual connections begin from Block 2. \\
\bottomrule
\end{tabular}
\end{center}
\end{table}

\begin{table}
\begin{center}
\captionof{table}{Description of Plain CNN used. }
\label{tab:plainCNN_desc}
\footnotesize
\begin{tabular}{c}
\toprule
\textbf{PlainCNN} \\
\midrule
Input: RGB Image ($32\times32$) \\
Conv $3\times3$, BN, Activation \\
$(L-1)\times$ Conv $3\times3$, BN, Activation \\
MaxPool at regular intervals (see released code) \\
Global Average Pooling \\
Linear Classifier \\
\bottomrule
\end{tabular}
\end{center}
\end{table}

\begin{table}
\begin{center}
\captionof{table}{Description of ResNet-18 and ResNet-50 used }
\label{tab:resnet1850_desc}
\footnotesize
\begin{tabular}{c}
\toprule
\textbf{ResNet-18 / ResNet-50} \\
\midrule
Input: RGB Image ($32\times32$) \\
$3\times3$ Conv Stem, BN, ReLU \\
Stage 1: Residual Blocks (64) \\
Stage 2: Residual Blocks (128, stride 2) \\
Stage 3: Residual Blocks (256, stride 2) \\
Stage 4: Residual Blocks (512, stride 2) \\
Global Average Pooling \\
Linear Classifier \\
\midrule
BasicBlock (ResNet-18), Bottleneck (ResNet-50) \\
\bottomrule
\end{tabular}
\end{center}
\end{table}

\begin{table}[t]
\centering
\caption{Description of Vision Transformer (ViT) used.}
\label{tab:vit_description}
\begin{tabular}{c}
\hline
\textbf{Vision Transformer (ViT)} \\
\hline
Input: RGB Image ($32 \times 32$) \\
Patch Embedding: Conv $4 \times 4$, Stride 4, Dim $d$ ($64$ Patches) \\
Prepend $[$CLS$]$ Token ($1 \times d$) + Positional Embedding ($65 \times d$) \\
$L \times$ Pre-LN Transformer Blocks ($L \in \{3, 6, 9\}$): \\
\quad Multi-Head Self-Attention ($h \in \{2, 4\}$, $d_k = 32$) $\pm$ Residual \\
\quad Transformer MLP ($d \to 2d \to d$, Activation) $\pm$ Residual \\
Final LayerNorm on $[$CLS$]$ Token (Penultimate Representation) \\
Linear Classifier ($d \to 10$) \\
\hline
\end{tabular}
\end{table}

\subsection{Experimental Sweeps Census}
Across the entire benchmark library, we evaluated $4{,}084$ synthetic models and $306$ real-world CIFAR-10 vision models (Total = 4,390 trained deep learning models) RTX 2080 Ti GPUs and PyTorch have been used for the experimentation purposes.
\subsubsection{Synthetic Regression Sweeps Library $M_\mathrm{models}=4{,}084$ Models} 
\textbf{Target Task - } Low-Rank Regression \\ 
Description of Synthetic Full Grid Sweep -Table~\ref{tab:synthetic_full_grid_sweep}\\
Description of Synthetic Recovery Grid Sweep - Table ~\ref{tab:synth_recovery_10k} \\
Description of Synthetic Residual Runs 1 - Table ~\ref{tab:synth_residual_sweep1} \\
Description of Synthetic Residual Runs 2 extended - Table ~\ref{tab:synth_residual_sweep_extended} \\

\paragraph{Synthetic data Full grid Sweep}
This experiment is conducted on a full grid of multiple architectures, optimizers, initialization schemes, task ranks, depths etc. To get a full generalized view of how a network's representation's dynamic changes as per different configurations.
\textbf{Total Runs - } $2,268$ , \textbf{Epochs - } $1000$ , Table~\ref{tab:synthetic_full_grid_sweep}

\begin{table}[H]
\centering
\small
\setlength{\tabcolsep}{6pt}

\fontsize{8.5}{9.5}\selectfont
\setlength{\tabcolsep}{4pt}
\captionof{table}{Synthetic Full Grid Sweep}
\label{tab:synthetic_full_grid_sweep}
\begin{tabular}{ll}
\toprule
\textbf{Common Settings} & \\
\midrule
Depth & $\{1,2,4,8,16,32,64\}$ \\
Task Rank & $\{1,4,16,24,32,64\}$ \\
Optimizer & SGD, Momentum SGD, Adam \\
Initialization & Default, Low-Rank \\
Runs  & 2268 (total) \\
\midrule
\textbf{Architecture} & \textbf{Configuration} \\
\midrule

Deep Linear ($252$ runs) &
Activation: Linear \\

Deep Residual MLP ($252$ runs) &
\begin{tabular}[t]{@{}l@{}}
Activation: ReLU\\
Residual: Pre-activation
\end{tabular} \\

Plain MLP ($1764$ runs) &
\begin{tabular}[t]{@{}l@{}}
Activations: Linear, ReLU, Tanh,\\
Leaky ReLU, SELU, GELU, SIREN
\end{tabular} \\

\bottomrule
\end{tabular}
\end{table}

\paragraph{Synthetic Recovery grid sweep.} This particular configuration aims to check whether a longer training time has any effect on recovery or no. Task rank is kept fixed because task rank has no detectable effect on representation dynamics. \textbf{Epochs - } $10,000$. Table ~\ref{tab:synth_recovery_10k}

\begin{table}
\centering
\small
 \caption{Synthetic Recovery grid sweep.}\label{tab:synth_recovery_10k}

\begin{tabular}{ll}

\toprule
\textbf{Setting} & \textbf{Value} \\
\midrule
Architecture & Plain MLP \\
Runs & 24 \\
Activations & SIREN, Leaky ReLU \\
Depth & $\{2,4,8,16,32,64\}$ \\
Task Rank & 24 \\
Optimizers & SGD, Adam \\
Initialization & Low-Rank \\
\bottomrule
\end{tabular}
\end{table}

\paragraph{Synthetic Residual Runs 1.} Experiment to check the effect of different residual implementation strategies on representation dynamics. \textbf{Epochs - }1000. Table ~\ref{tab:synth_residual_sweep1}

\begin{table}[H]
\centering
\small
 \caption{Synthetic Residual Runs Sweep }\label{tab:synth_residual_sweep1}
\begin{tabular}{ll}
\toprule
\textbf{Setting} & \textbf{Value} \\
\midrule
Architecture & Residual MLP \\
Runs & 448 \\
Depth & $\{1,2,4,8,16,32,64\}$ \\
Activations & SIREN, Leaky ReLU \\
Residual Types & Post-activation, Depth-scaled, \\
& Fixed-$0.5$, Learnable Gate \\
Task Ranks & $\{1,64\}$ \\
Optimizers & SGD, Adam \\
Initialization & Default, Low-Rank \\
\bottomrule
\end{tabular}
\end{table}

\paragraph{Synthetic Residual Runs 2.} An extended configuration runs of Synthetic Residual Runs 1. \textbf{Epochs - }1000. Table ~\ref{tab:synth_residual_sweep_extended}

\begin{table}[H]
\begin{center}
\small
 \caption{Synthetic Residual sweep extended.}\label{tab:synth_residual_sweep_extended}
\begin{tabular}{ll}
\toprule
\textbf{Setting} & \textbf{Value} \\
\midrule
Architecture & Residual MLP \\
Runs & 1344 \\
Depth & $\{2,4,8,16,32,64\}$ \\
Task Rank & 24 \\
Activations & Linear, ReLU, Tanh, Leaky ReLU, \\
& SELU, GELU, SIREN \\
Residual Types & Post-activation, Pre-activation, \\
& Depth-scaled, Fixed-$0.1$, Fixed-$0.25$, \\
& Fixed-$0.5$, Fixed-$0.75$, Learnable Gate \\
Optimizers & Adam, Momentum SGD \\
Initialization & Default, Low-Rank \\
\bottomrule
\end{tabular}
\end{center}  
\end{table}

\subsubsection{Real World Dataset Sweep}
Description of CNN, Plain MLP, ResMLP sweep on CIFAR10 - Table ~\ref{tab:cifar10_cnn_resmlp_mlp} \\
Description of CNN only on CIFAR10 sweep - Table ~\ref{tab:cnn_settings_192} \\
Description of ResNet18 , ResNet50 on CIFAR10 sweep - Table ~\ref{tab:resnet1850cifar10sweep} \\
Description of ViT sweep - Table~\ref{tab:vit_grid_sweep} \\
\paragraph{CIFAR10 CNN, MLP, ResMLP sweep.} This sweep's aim is to observe how representation dynamics appear across models with different settings on Real world dataset such as CIFAR10. This also checks whether effect of width of a layer and input data sample size. \textbf{Total Runs - }60, \textbf{Epochs - }200

\begin{table}[H]
\begin{center}
\small
 \caption{CIFAR10 CNN, Plain MLPs, Residual MLPs sweep}\label{tab:cifar10_cnn_resmlp_mlp}
\begin{tabular}{ll}
\toprule
\textbf{Setting} & \textbf{Value} \\
\midrule
Architectures & MLP, Residual MLP, CNN \\
Expansion Factor & $\{1,2,4,8\}$ \\
Samples/Class & Full, 20, 50, 100, 200 \\
\bottomrule
\end{tabular}
\end{center}  
\end{table}

\paragraph{CNN only CIFAR10 sweep.} This experiment's aim is to observe how representation dynamics work in Plain CNN configurations. \textbf{Total Runs - }192, \textbf{Epochs - }300.

\begin{table}[H]
\centering
\caption{CNN only on CIFAR10 sweep}
\label{tab:cnn_settings_192}

\small
\begin{tabular}{ll}
\toprule
\textbf{Setting} & \textbf{Value} \\
\midrule
Architecture & Plain CNN \\
Depth & $\{8,32\}$ \\
Activations & Leaky ReLU, SIREN \\
Expansion Factor & $\{1,2,4\}$ \\
Optimizers & SGD, Adam \\
Initialization & Default, Low-Rank \\
Samples/Class & 20, 100, 1000, Full \\
\bottomrule
\end{tabular}
\end{table}

\paragraph{ResNet CIFAR10 sweep.} The motivation of this experiment is to observe how representation dynamics operate in standard ResNet-18 and ResNet-50 with 3 different residual connection implementation styles. \textbf{Total Runs - }6, \textbf{Epochs}-2500.
\begin{table}[H]
\begin{center}
 \caption{ResNet CIFAR10 sweeps}\label{tab:resnet1850cifar10sweep}
\small
\begin{tabular}{ll}
\toprule
\textbf{Setting} & \textbf{Value} \\
\midrule
Architecture & ResNet \\
Models & ResNet-18, ResNet-50 \\
Residual Skip Modes &
\begin{tabular}[t]{@{}l@{}}
On: $x+F(x)$\\
Off: $F(x)$\\
Scaled: $x+\frac{1}{N_{blocks}}F(x)$
\end{tabular} \\
\bottomrule
\end{tabular}
\end{center}  
\end{table}
\textbf{ViT CIFAR10 sweep.} The motivation of this experiment is to observe how representation dynamics operate in Vision Transformers across 3 depths and 2 widths with standard and non-residual connection styles. \textbf{Total Runs - 48 , Epochs - 200}
\begin{table}[H]
\centering
\caption{Factorial Grid Sweep Configurations for Vision Transformers ($N=48$ Models).}
\label{tab:vit_grid_sweep}
\begin{tabular}{llc}
\hline
\textbf{Hyperparameter} & \textbf{Values / Levels} & \textbf{Count} \\
\hline
Residual Mode & Standard (with residuals), No-Residual (without residuals) & 2 \\
Depth ($L$) & 3, 6, 9 blocks & 3 \\
Embedding Dim ($d$) & 64 ($h=2, d_{\text{mlp}}=128$), 128 ($h=4, d_{\text{mlp}}=256$) & 2 \\
Activation ($\Phi$) & GELU, ReLU & 2 \\
Optimizer & Adam ($\eta=10^{-3}$), Momentum SGD ($\eta=10^{-2}, \beta=0.9$) & 2 \\
\hline
\textbf{Total Configurations} & $2 \times 3 \times 2 \times 2 \times 2$ & \textbf{48} \\
\hline
\end{tabular}
\end{table}

\end{document}